\documentclass[preprint,12pt,authoryear,times]{elsarticle}

\usepackage{amssymb}
\usepackage{lineno}
\usepackage{booktabs}
\usepackage{threeparttable}
\usepackage{caption}
\usepackage{subcaption}
\usepackage{url}
\usepackage{amsmath} 
\usepackage{graphicx}
\usepackage{array}
\usepackage{booktabs}
\usepackage{multirow}
\usepackage{rotating}
\usepackage{makecell}
\usepackage{hyperref}
\usepackage[table]{xcolor}  
\usepackage{adjustbox}  

\hypersetup{hypertex=true,
    colorlinks=true,
    linkcolor=true,
    anchorcolor=true,
    citecolor=true}

\journal{}

\begin{document}

\begin{frontmatter}

\title{From Street View to Visual Network: \\Mapping the Visibility of Urban Landmarks with Vision–Language Models}

\author[doa]{Zicheng Fan}
\author[doa,tak]{Kunihiko Fujiwara}
\author[uog]{Pengyuan Liu}
\author[pku]{Fan Zhang}
\author[doa,dre]{Filip Biljecki\corref{cor1}}

\cortext[cor1]{Corresponding author.}

\affiliation[doa]{organization={Department of Architecture, National University of Singapore}, country={Singapore}}
\affiliation[tak]{organization={Research \& Development Institute, Takenaka Corporation}, country={Japan}}
\affiliation[uog]{organization={Urban Analytics Subject Group, Urban Studies \& Social Policy Division, University of Glasgow}, country={United Kingdom}}
\affiliation[pku]{organization={Institute of Remote Sensing and GIS, Peking University}, country={China}}
\affiliation[dre]{organization={Department of Real Estate, National University of Singapore}, country={Singapore}}

\begin{abstract} 

Visibility analysis in urban planning has traditionally relied on line-of-sight (LoS) simulations, which capture geometric occlusion. However, these approaches depend on accurate 3D data that is often unavailable and may not adequately represent how visually distinctive urban landmarks are encountered in real streetscapes.
We reformulate landmark visibility assessment as an urban visual search problem in image space by leveraging the widespread availability of street view imagery (SVI). Given a reference image of a target landmark, a Vision Language Model (VLM) is applied to detect the landmark in direction- and zoom-controlled SVI. A successful detection indicates machine-recognised landmark visibility at the corresponding viewpoint.
Beyond isolated viewpoints, we construct a heterogeneous visibility graph to represent visual connectivity among landmarks, street-view locations, and the urban spaces that mediate them. This graph enables us to map where visual connections occur, how strong they are, and how multiple landmarks become jointly connected through shared visual corridors.
Across six well-known landmark structures in global cities, the image-based method achieves an overall detection accuracy of 87\%, with a precision score of 68\% for landmark-visible locations. In a second case study along the River Thames in London, the visibility graph reveals multi-landmark connections and identifies key mediating locations, with bridges accounting for approximately 31\% of all connections.
The proposed method complements LoS-based visibility analysis and offers a practical alternative in data-constrained settings. It also showcases the possibility of revealing the prevalent connections of visual objects in the urban environment, opening new perspectives for urban planning and heritage conservation.

\end{abstract}


\end{frontmatter}

\section{Introduction}
\label{sec:sample1}

Evaluating the visibility of key landmarks and the visual impact of new obstructions is of critical importance for urban planning, especially in heritage conservation and development control \citep{lopes_assessment_2019, czynska_classification_2019, ashrafi_heritage_2021, talamini_visibility_2023}. Landmarks are closely tied to the identity and spirit of their cities and often serve as recognisable visual symbols \citep{bruns_influence_2019,manahasa_role_2024}. Protecting the visibility of historic landmarks and their visual relationships with surrounding urban spaces, for example, through visual corridors \citep{greater_london_authority_london_2012}, can help cities maintain historical continuity and preserve a distinctive image of the city \citep{lynch_image_1996}. Meanwhile, assessing and regulating the visual impact of new buildings, especially high-rise developments, can support a more coherent urban skyline and help protect valued natural landscapes \citep{city_of_vancouver_public_2024}.

Visibility analysis is traditionally conducted based on the Line of Sight (LoS) simulation \citep{fisher-gewirtzman_voxel_2013, cilliers_critical_2023}.
LoS establishes a hypothetical direct line between an observer and the target object, with the presence of an uninterrupted line indicating visibility. Owing to its simplicity and intuitive nature, LoS has been widely adopted and underpins more advanced visibility analyses, such as 2D and 3D viewshed analysis \citep{amidon_delineating_1968,kloucek_how_2015,inglis_viewsheds_2022}, isovist analysis \citep{tandy_isovist_1967, benedikt_take_1979, batty_exploring_2001}, to map the extent and boundaries of visible space, and visibility graph analysis to investigate the spatial relationship defined by visual links \citep{turner_isovists_2001}. %

\begin{figure}[tbp]
    \centering
    \makebox[\textwidth][c]{%
        \includegraphics[width=1\textwidth]{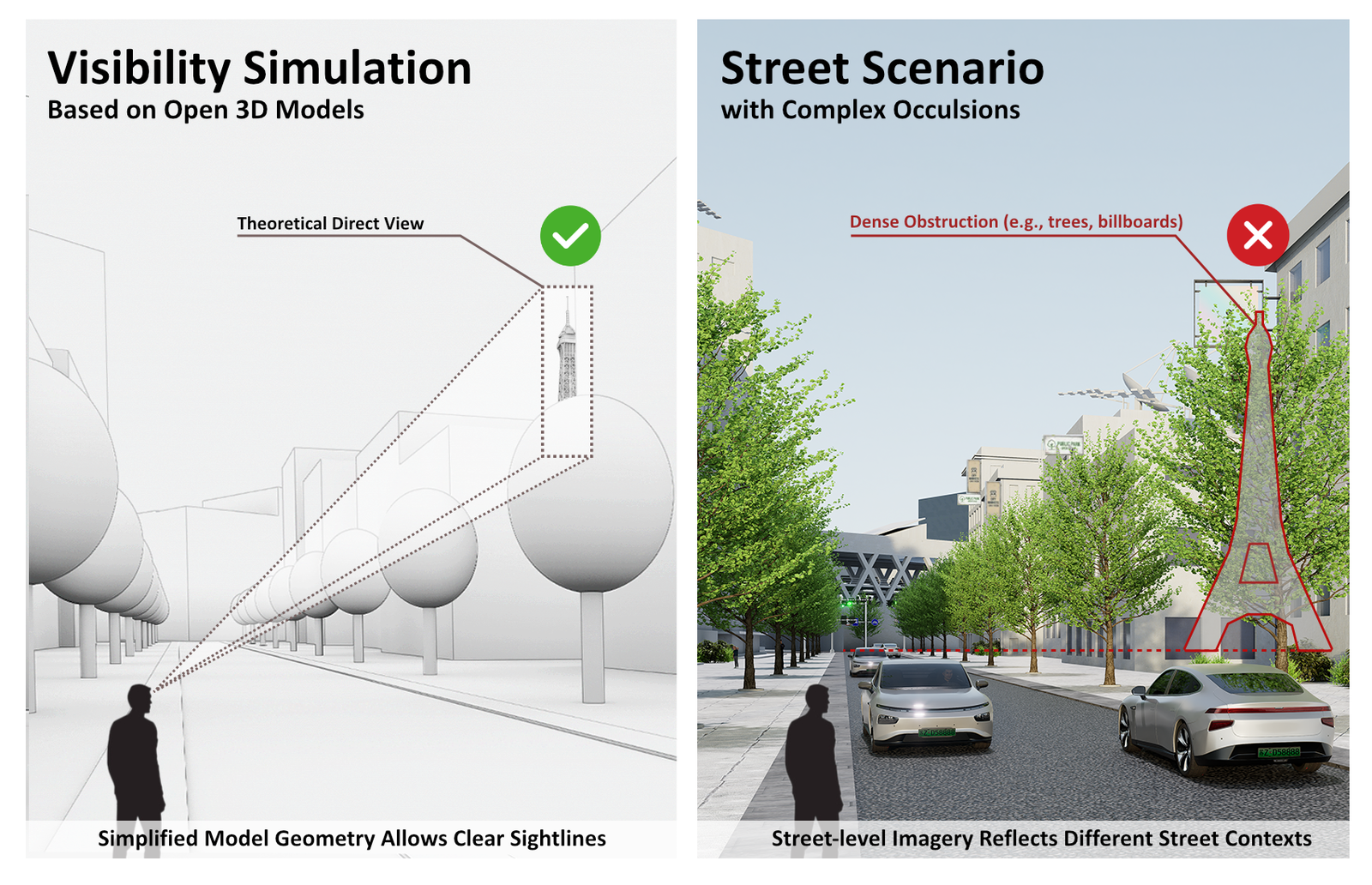}%
    }
    \caption{Complex street environment, especially the existence of various occlusions, such as trees and billboards, makes it possible for overestimating or underestimating landmarks' visibility via line-of-sight simulation, but also provides opportunities for image-based methods.}
    \label{fig:concept}
\end{figure}

Despite the widespread use, visibility analysis based on LoS faces two key limitations when applied in the 3D urban environment: 
First, its reliability relies on the completeness and resolution of the underlying spatial data models, such as building footprints with associated height attributes, 3D city models, Digital Surface Models (DSM), and Digital Elevation Models (DEM) \citep{wrozynski_reaching_2024, cilliers_critical_2023,morello_digital_2009,lei_assessing_2023}. However, high-quality, complete, and openly available 3D datasets remain scarce in many cities. As a result, some studies continue to rely on 10--30 m DEMs, whose coarse resolution can misrepresent LoS relationships and introduce substantial errors into visibility estimation \citep{inglis_viewsheds_2022}. This limitation becomes more evident when analysing named objects such as urban landmarks at large spatial scales, where both 3D data quality and spatial coverage are often insufficient.

Second, traditional LoS‑based methods can be blind to visual context and inter‑object relationships, %
which are important to how urban scenes are visually recognised and interpreted \citep{gillings_seeing_2001,dederix_patterns_2019,inglis_viewsheds_2022}.
As shown in Figure \ref{fig:concept}, by treating visibility as a geometric intersection, LoS methods based on simplified 3D models may overlook the contextual factors, such as lighting, vegetation, advertising boards, architectural embellishments, and the way multiple objects share a view.
Tracing an unobstructed LoS to the apex of a landmark may confirm it is “visible”, but this says little about how the landmark is framed by neighbouring buildings, how much of its recognisable silhouette emerges beyond tree canopies, or how it co‑occurs with other icons along a skyline.
Additionally, pedestrians' visual experience of landmarks occurs not only at fixed locations, such as tourism viewpoints, but also during everyday movement along the road network. In this context, visibility is associated not merely with points but also with extended spatial configurations such as paths, corridors, and regions, suggesting the need for a more systematic, network-based thinking of visibility.
Without incorporating such contextual cues and inter‑object relationships, LoS analysis may not fully reflect the visually recognisable conditions encountered in real streetscapes.

\begin{figure}[htbp]
    \centering
    \makebox[\textwidth][c]{%
        \includegraphics[width=1.2\textwidth]{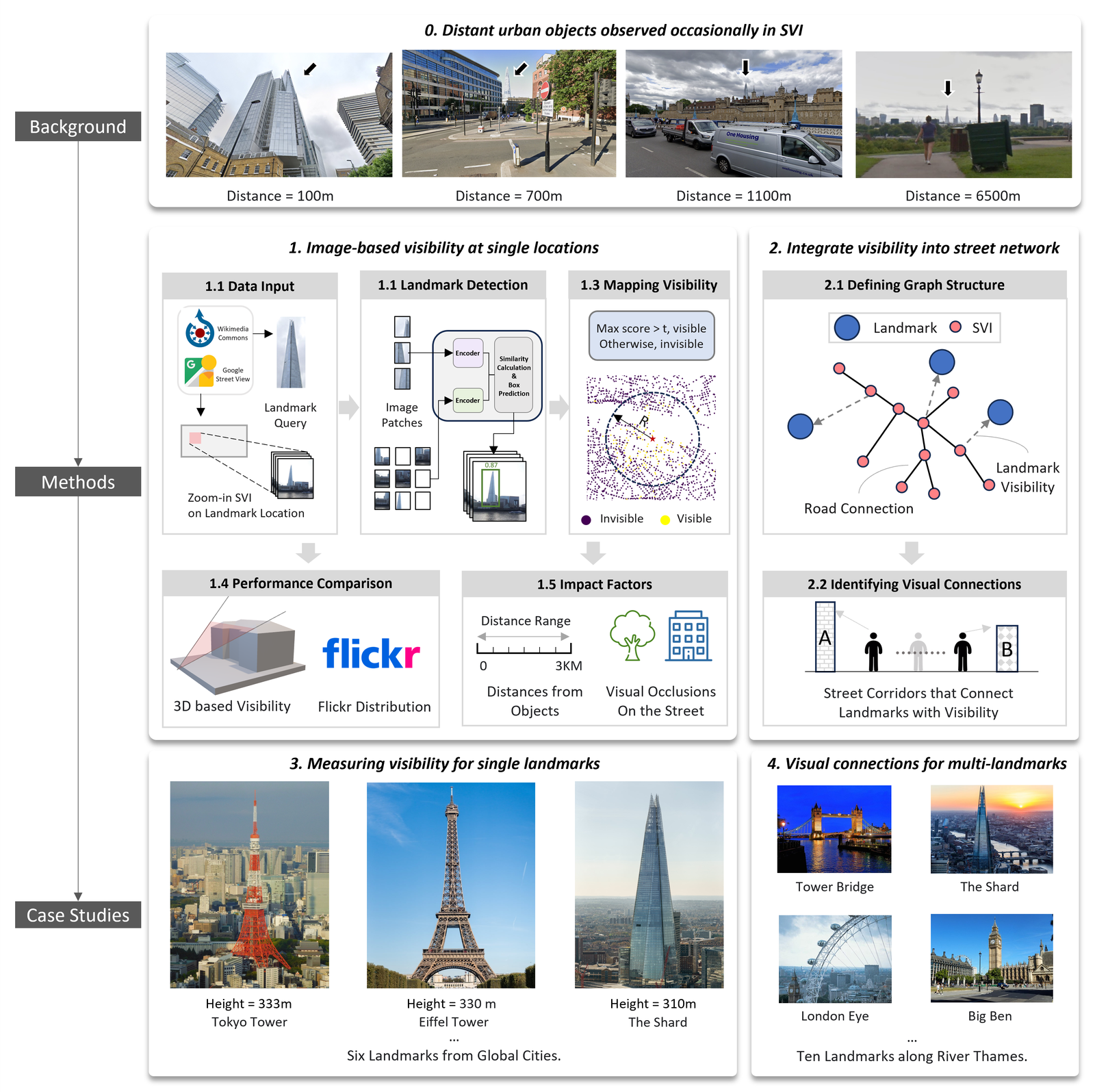}%
    }
    \caption{A research framework for the study. Imagery: Google Street View, Wikimedia Commons, Tripadvisor.}
    \label{fig:concept_landmark}
\end{figure}

Unlike LoS analysis, which simulates the visual process based on data models, images record the outcome of visibility in place, capturing real-world visual relationships directly.
In previous practices, images %
often serve as supplementary reference to LoS-based visibility analysis, offering intuitive insights about both visibility and visual contexts \citep{greater_london_authority_london_2012, talamini_visibility_2023}. However, their potential for large-scale visibility assessment of recognisable urban landmarks remains underexplored. 
Recent advancements in
Vision-Language Models (VLMs), such as CLIP \citep{radford_learning_2021}, OWL-ViT \citep{minderer_simple_2022,minderer_scaling_2024}, Grounding DINO \citep{liu_grounding_2024} 
make it increasingly feasible to identify named landmarks in images by their distinctive visual characteristics, such as unique height, architectural style, or silhouette.
Meantime, Street View Imagery (SVI) has emerged as an important data source that provides dense and spatially referenced viewpoints along urban streets \citep{ito_understanding_2024}. Unlike isolated photographs, SVI %
allows image-based visibility inference to be scaled up geographically.
SVI is typically used to observe and map the foreground space of a city for less visual obstruction \citep{fan_coverage_2025,liang_evaluating_2024}, while neglecting distant objects in the background. 
However, as shown in Figure \ref{fig:concept_landmark}, 
certain iconic urban landmarks (e.g., The Shard in London) can be occasionally observed from SVI at street level, from varying distances and urban settings. 
These observations enable landmark visibility to be assessed at individual viewpoints, while the spatial continuity among SVI locations further allows such point-based records to be connected into a city-scale visual network. 
This extension has great potential in revealing broader visual relationships and spatial configurations, complementing existing heritage conservation strategies and supporting more actionable policies for urban planning and management.

Given these gaps and opportunities, this study explores an image-based visibility detection method that leverages computer vision and geo-located SVI, offering a data-driven complement to conventional LoS-based visibility analysis. The questions addressed in this study include:
    
    \begin{enumerate}
        \item Can images be quantitatively applied as a scalable approach to investigate the visibility of recognisable urban landmarks, especially where detailed 3D models are unavailable?

        \item To what extent can images reveal the visual relationships among multiple landmarks and their interaction with the urban environment, and support urban planning applications?

    \end{enumerate}

A research framework is presented in Figure \ref{fig:concept_landmark}. Firstly, we introduce the workflow %
for detecting the visibility of distant landmarks via SVI. Using a case study, we assess the method's reliability in evaluating the visibility of the iconic high-rise landmarks across global cities. We also show its unique value in addressing diverse visual observation contexts compared with the open 3D setup used in this study. 
We then extend the application of image-based visibility to analysing visual relationships among multiple landmarks by developing a heterogeneous visibility graph. Using the second case study, we examine broader visual-spatial connections among multiple landmarks along the River Thames in London, UK. The advantages, limitations, and practical applications of the proposed method are discussed accordingly.

\section{Background and Related Work}

\subsection{Common Methods and Limitations for Investigating Visibility}

\subsubsection{2D and 3D Visibility Analysis}
Visibility analysis, based on the Line of Sight (LoS) principle, is fundamental in urban planning and landscape research, and has been widely applied to investigate the visual condition and impact of urban landmarks \citep{santosa_visibility_2023, pyka_lidar-based_2022,mor_3d_2021,czynska_classification_2019,bartie_identifying_2015}. Historically, 2D visibility analysis emerged first and gained wide application in architectural and urban design fields. By assuming uniform ground elevation and treating environmental obstructions (e.g., walls, buildings, trees) as binary barriers, 2D LoS simplifies the definition and identification of visibility. This simplification enabled early exploration of visual relationships and spatial configurations. A classical method is isovist analysis, introduced by \citet{tandy_isovist_1967} and refined by \citet{benedikt_take_1979}, which defines and describes the immediate visible area from a given observer location. Isovist-based 2D methods are widely used to evaluate building layouts \citep{batty_exploring_2001, hosseini_alamdari_new_2022} and investigate human perception and behaviour in the space \citep{wiener_isovists_2005,krukar_embodied_2021,snopkova_isovists_2023}. However, 2D methods often rely
on oversimplified assumptions about ground uniformity and remove metric information \citep{ratti_space_2004}, constraining their applicability in complex terrains and vertical morphologies. 

To overcome these constraints, 3D visibility analysis evolved to simulate real-world scenes more accurately. Several studies extended isovist analysis to 3D spatial environments \citep {hagberg_exploring_2008,kim_new_2019,krukar_embodied_2021}.
In parallel, 3D viewshed analysis has become common practice in large-scale applications such as evaluating landscape visual quality \citep{czynska_classification_2019,swietek_visual_2023} or assessing the visual impact of new developments \citep{inglis_viewsheds_2022, kloucek_how_2015, alphan_modelling_2021, cilliers_critical_2023}. 
2.5D data models, such as the Digital Surface Models (DSM), Digital Elevation Models (DEM), and vector-based building data, are applied to simulate the real-world environment surfaces \citep{czynska_classification_2019}. 
Viewshed, cumulative viewshed, and fuzzy viewshed analyses \citep{cilliers_critical_2023} are now standard tools in GIS platforms (e.g., ArcGIS \citep{esri_visibility_2021}, QGIS \citep{cuckovic_advanced_2016}, GRASS \citep{neteler_grass_2012}).

Despite their enhanced realism, 
2.5D models frequently lack semantic differentiation and are restricted to representing complex 3D morphology \citep{pyka_lidar-based_2022}. To address this, recent studies apply true 3D data, such as LiDAR point clouds and voxel grids, and ray-tracing methods, to enhance detail and accuracy \citep{kim_new_2019,wu_mapping_2021,zhao_mapping_2020,wrozynski_reaching_2024, fujiwara_voxcity_2026}. 
For instance, \citet{wrozynski_reaching_2024} combined GIS with 3D graphic software and uses high-resolution LIDAR data to quantify the landscape impacts of
 photovoltaic farms, where a better representation of vegetation was achieved.
These developments mark a shift from simplified surface-based visibility modelling towards more detailed geometric simulation of urban visual fields. Even so, their large-scale implementation is also computationally intensive.

With advances in 3D geovisualization techniques, photorealistic 3D city models and rendering-based approaches were also explored to analyze visibility and visual exposure of urban elements \citep{li_room_2022, buyukdemircioglu_generating_2025, park_window_2026, liu_linking_2026}. Compared with purely geometric 3D visibility methods, rendering-based approaches can incorporate not only the geometry of urban objects but also image textures, facade appearances, and viewpoint-dependent visual presentation.
Nevertheless, these methods still depend heavily on the availability, completeness, and quality of detailed 3D city models and texture information, especially when applied across multiple cities.

\subsubsection{Graph-Based Extensions of Visibility Analysis}

Graph‐based methods extend visibility analysis beyond isolated viewpoints to examine visual connectivity among multiple locations. Visibility Graph Analysis (VGA), rooted in space syntax theory \citep{hillier_social_1984, hillier_space_1996}, models environments as 2D grids where nodes connect if their corresponding spaces are mutually visible \citep{turner_isovists_2001}. Metrics like mean depth and centrality describe spatial configuration and predict human movement patterns. While VGA is effective for structured indoor settings or defined outdoor areas, it inherits limitations from LoS-based methods: vertical complexity and environmental semantics are often oversimplified. To address these gaps, efforts continue to develop 3D visibility graphs \citep{varoudis_beyond_2014,lu_three-dimensional_2019,omrani_azizabad_three-dimensional_2024}. For semantic enrichment, the Integrative Visibility Graph (IVG) incorporates functional nodes (e.g., food and drink facilities) \citep{natapov_can_2013}, while the Functional Visibility Graph (FVG) systematically links urban functions to spatial nodes via visibility edges, measuring visual accessibility to specific activities \citep{shen_functional_2022}. Despite these advancements, existing graph-based methods remain grounded in geometric LoS simulations when defining the visibility, %
which may not fully capture street-level visual context.
In contrast, \citet{bartie_identifying_2015} introduces a novel approach to identifying landmark relationships from single images based on semantic similarity, presenting the potential to %
study landmark visibility and visual association from image data.

\subsection{Street View Images as Proxy for Visibility Analysis}
\subsubsection{Visibility Analysis Centring on Semantic Elements}

The emergence of SVI as an essential data source in urban studies has expanded the scope of visibility analysis towards semantic elements and instances.
Commonly collected along road networks by map service providers, SVI presents good organisation and availability in metadata, such as coordinates, heading and field of view \citep{anguelov_google_2010,hou_comprehensive_2022}. 
With the convenience, SVI is commonly used as a street-level visual proxy in urban environments \citep{ito_understanding_2024}.
Relying on Computer Vision (CV) models, researchers can easily detect the pixels of buildings, trees, or designated object instances from SVI and calculate their existence ratio in the image frame. The pixel ratio, as a visibility metric, represents the extent to which the potential observer in the street environment perceives the element or instances.
The Green View Factor (GVF), for example, is a metric of the perceived greenery element in streetscape, and is proven relevant to property price \citep{yang_financial_2021,xu_evaluation_2025}, and mental health conditions \citep{belcher_socioeconomic_2024}. Similarly, Building View Factor (BVF) and Sky View Factor (SVF) derived from SVI have become important metrics for describing urban canyon \citep{hu_classification_2020} from a human perspective, and play roles in Local Climate Zone (LCZ) classification \citep{li_fine-grained_2025} and climate modelling \citep{middel_sky_2018, Fujiwara2024-vs}.

Common use cases detect visual elements or instances in the foreground of SVI, as they are close enough to the camera, providing sufficient colour or texture details for CV recognition. However, background pixels are often dropped and ignored. Elements appearing in the background are usually blocked by the foreground buildings or trees \citep{fan_coverage_2025}, or presented in low resolution, adding difficulty for machine detection. Nevertheless, it is argued that valuable messages are naturally embedded in whether urban objects are visible or not in the image background. 
As a typical example, urban landmarks often appear in the background of a pedestrian’s field of view, and their extent of presence is related to the attractiveness of the observation location or path \citep{mor_3d_2021}. \citet{pyka_lidar-based_2022} used high-resolution LiDAR data to generate synthetic panoramas and evaluated the visual exposure of landmarks as background elements, demonstrating the potential of image mediums for landmark visibility analysis.

\subsubsection{Imaging Lines of Sight via Visual-Language Models}
The development of Vision-Language Models (VLM) %
has lowered the barriers to acquiring and interpreting urban information from visual data. 
VLMs enable correspondences to be established between image data, such as satellite imagery and SVI and real-world semantics.
Contrastive Language-Image Pre-training model (CLIP) \citep{openai_clip_2022}, developed by OpenAI, is a typical example of VLM. The model takes images and their corresponding textual descriptions as input and learn to associate the knowledge from both modalities by computing the similarity between text and image embeddings and minimising contrastive loss \citep{radford_learning_2021}. Applying the principles of CLIP, \citet{klemmer_satclip_2024} develops the `SatCLIP' for matching geographic coordinates and visual characteristics extracted from satellite imagery. \citet{huang_zero-shot_2024} propose the `UrbanCLIP', which infers urban functions using SVI and a fine-tuned CLIP model. 

A more prevalent application of VLM is zero-shot object detection, which allows searching and localising target objects in image space without additional pre-training or fine-tuning of the model. Well-known zero-shot object detection models include Grounding DINO \citep{liu_grounding_2024} and OWL-ViT \citep{minderer_simple_2022, minderer_scaling_2024}. Both of them accept textual queries for searching the target, while OWL also allows image queries. %
In this study, we focus on urban landmarks as visually distinctive targets that may be identified from SVI using zero-shot detection models. The detection process can be regarded as machine observation at a specific SVI location. 
In this way, landmark visibility and related visual context can be identified from image data, forming the basis of the image-based visibility analysis method. The details of the method will be elaborated in the following sections.

\section{Methodology}

\subsection{An Image-based Landmark Visibility Detection Method}
\label{sec:image_based_visibility}

\begin{figure}[tbp]
    \centering
    \makebox[\textwidth][c]{%
        \includegraphics[width=1.2\textwidth]{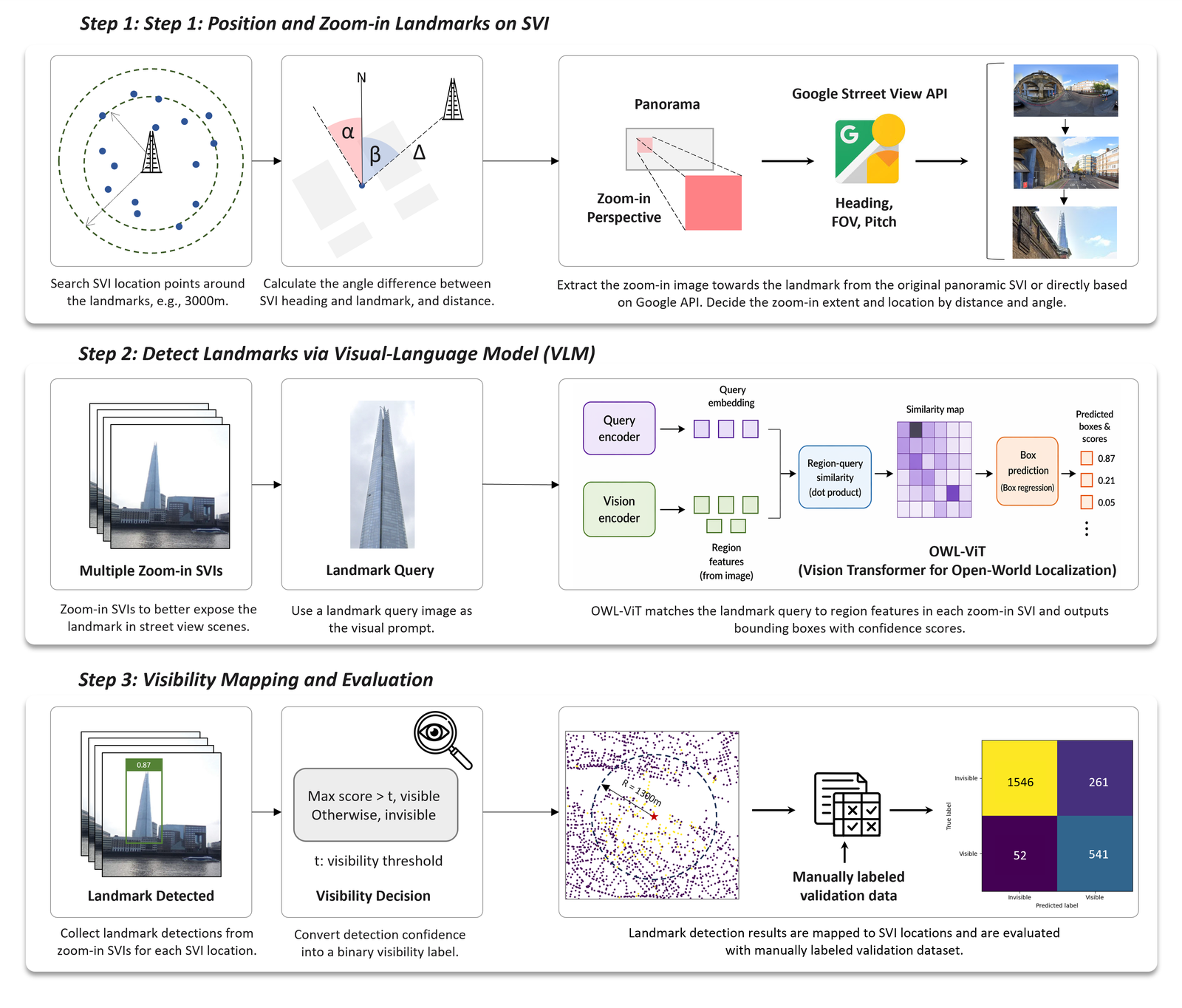}%
    }
    \caption{Workflow of locating and detecting the visibility of a distant landmark from SVI. Imagery: Google Street View, Wikimedia Commons.}
    \label{fig:framework_landmark}
\end{figure}

Taking the famous London landmark, The Shard, as an example, Figure \ref{fig:framework_landmark} illustrates the workflow for assessing and mapping its visibility from panoramic SVIs. Key steps include relative positioning and image zooming, object detection, and mapping and evaluation. Panoramic SVIs and metadata, including heading and coordinates, are retrieved from the Google Street View service.

\subsubsection{Step 1: Position and Zoom-in Landmark on SVI}
The process begins by identifying SVI location points within a set distance (e.g., 3000m) around a landmark as potential observers for the analysis. 
This radius is adopted here as an operational search extent for tall landmarks in dense urban settings.
Once these SVI locations are selected, the distances and relative angles between them and the landmark are calculated, which help position the landmark in image space. The panoramic SVI is then zoomed in towards the upper half of the landmark, which is commonly less blocked by other buildings. In this way, the landmark occupies a larger portion of the frame. This step also reduces the noise and complexity introduced by other visual elements that could affect the performance of the following object detection. The zoom-in extent is determined based on the distance between the SVI location and the landmark and the landmark's physical height. The detailed steps for locating and zooming in landmarks on SVI are presented in \ref{sec:landmark_position}.

\begin{figure}[tbp]
    \centering
    \begin{subfigure}{1\textwidth}
        \centering
        \includegraphics[width=\textwidth]{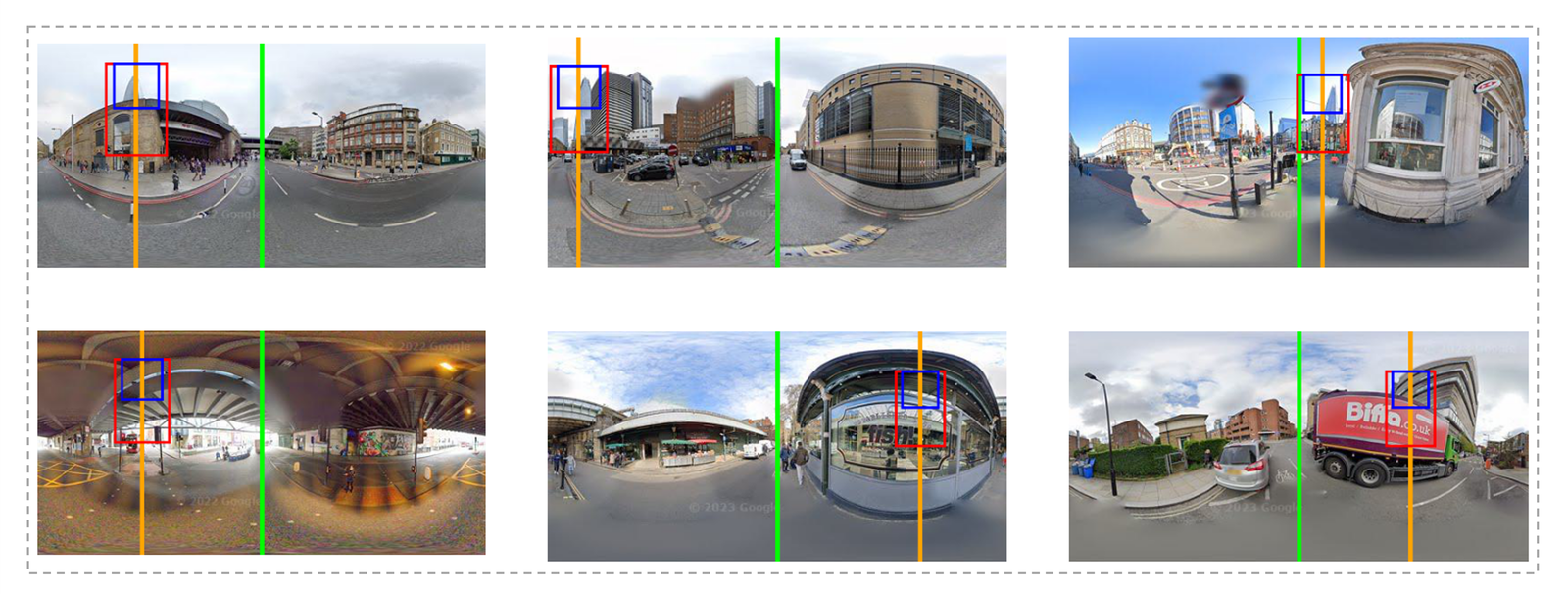}
        \caption{A showcase of localisation boxes with landmark visible (first row) and blocked (second row). Imagery: Google Street View.}
        \label{fig:zoom_in}
    \end{subfigure}
    
    \vspace{0.5cm} %
    
    \begin{subfigure}{1\textwidth}
        \centering
        \includegraphics[width=\textwidth]{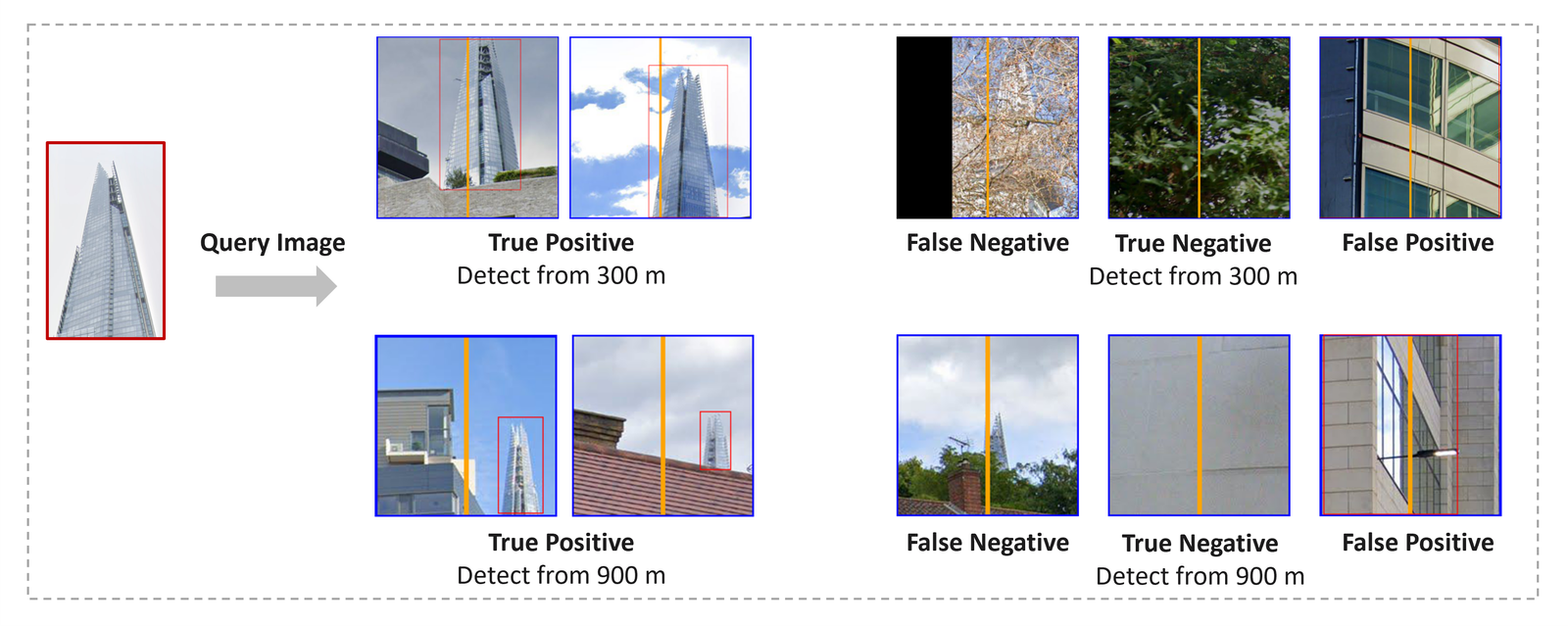}
        \caption{Example results of true positive, true negative, false positive and false negative from landmark detection. Imagery: Google Street View, Wikimedia Commons.}
        \label{fig:detect_landmark}
    \end{subfigure}
    
    \caption{Visualisation of the landmark detection process. (a) Locating landmarks using bounding boxes. (b) Detecting the landmark within the zoomed-in region.}
    \label{fig:combined_landmark}
\end{figure}

Figure \ref{fig:zoom_in} illustrates the attempts to locate The Shard, the highest building in the UK, in the image space of SVIs. The green line represents the heading of the camera, the yellow line represents the direction of The Shard relative to the SVI location, and the red and blue bounding boxes represent the areas where the whole and the upper half of the building may appear in the image, respectively. It is observed that The Shard can be occasionally obstructed by other buildings in SVIs.

\subsubsection{Step 2: Detect Landmarks via Visual-Language Model (VLM)}
The study conducts zero-shot object detection on the zoomed-in SVIs to identify potential landmarks. The task is supported by Vision Transformer for Open-World Localisation (OWL-ViT) \citep{minderer_simple_2022}, a model combining CLIP and lightweight object classification and localisation heads.

Specifically, the study adopts the image-guided method supported by the model for landmark detection. 
A query image representing the recognisable visual features of the landmark is input to find the most similar visual object on the zoomed-in image. The detection process generates confidence scores that indicate the likelihood of the landmark's presence in the images. In our implementation, image queries are more reliable than text queries, such as landmark names or shape descriptions, especially when landmarks are partially obstructed in complex street environments.

\subsubsection{Step 3: Visibility Mapping and Evaluation}
To evaluate the performance of the image-based visibility detection method, a validation dataset is generated for each landmark by dividing the total SVI locations into four distance bands from the central landmark and randomly sampling 100 locations per distance band. The sampled SVIs are manually audited and labelled as `visible' or `invisible' with respect to the target landmark, using the open-source labelling platform Label Studio \citep{tkachenko_label_2020}. Given that landmarks are generally distinctive in their overall shape, colour, texture, and architectural details, partial visibility is considered sufficient for a visible' label only when the landmark remains recognisable. Conversely, fragments that do not provide sufficient visual evidence for recognising the landmark identity are labelled as invisible'.
Figure \ref{fig:detect_landmark} presents example results of detecting landmarks with OWL-ViT model. It shows that with the zoom-in procedure, the model can recognise The Shard building from 300m and 900m, with the proper zoom-in processing. Even so, different kinds of classification mistakes can still occur. For example, the model may regard irrelevant buildings or components as The Shard itself. In other cases, The Shard is present in the image but heavily blocked by foreground trees, leading to potential false negatives.

\subsection{From Isolated Visibility to Visibility Network}
\label{sec:graph_construction}

Compared with LoS simulation based on 3D data, the image-based method also has its own data availability constraints. Specifically, SVIs are systematically collected along urban roads and present limited coverage within street blocks \citep{fan_coverage_2025}. In addition, the distribution of SVI within local urban street spaces may be uneven. Some roads may completely lack SVI coverage or be represented only by sparse viewpoints. These limitations mean that SVI-based visibility analysis cannot densely cover all available spatial units in the same way as LoS simulation. Nevertheless, because of its distinctive mode of data collection and production, SVI shows a natural affinity with urban road networks. This makes it possible to relate isolated visibility detections to paths or movement trajectories within the network, and to examine how landmarks become connected through sequences of street-level viewpoints. This characteristic of SVI provides a new window for investigating landmark visibility and visual relationships.

\begin{figure}[htbp]
    \centering
    \begin{subfigure}{1\textwidth}
        \centering
        \includegraphics[width=\textwidth]{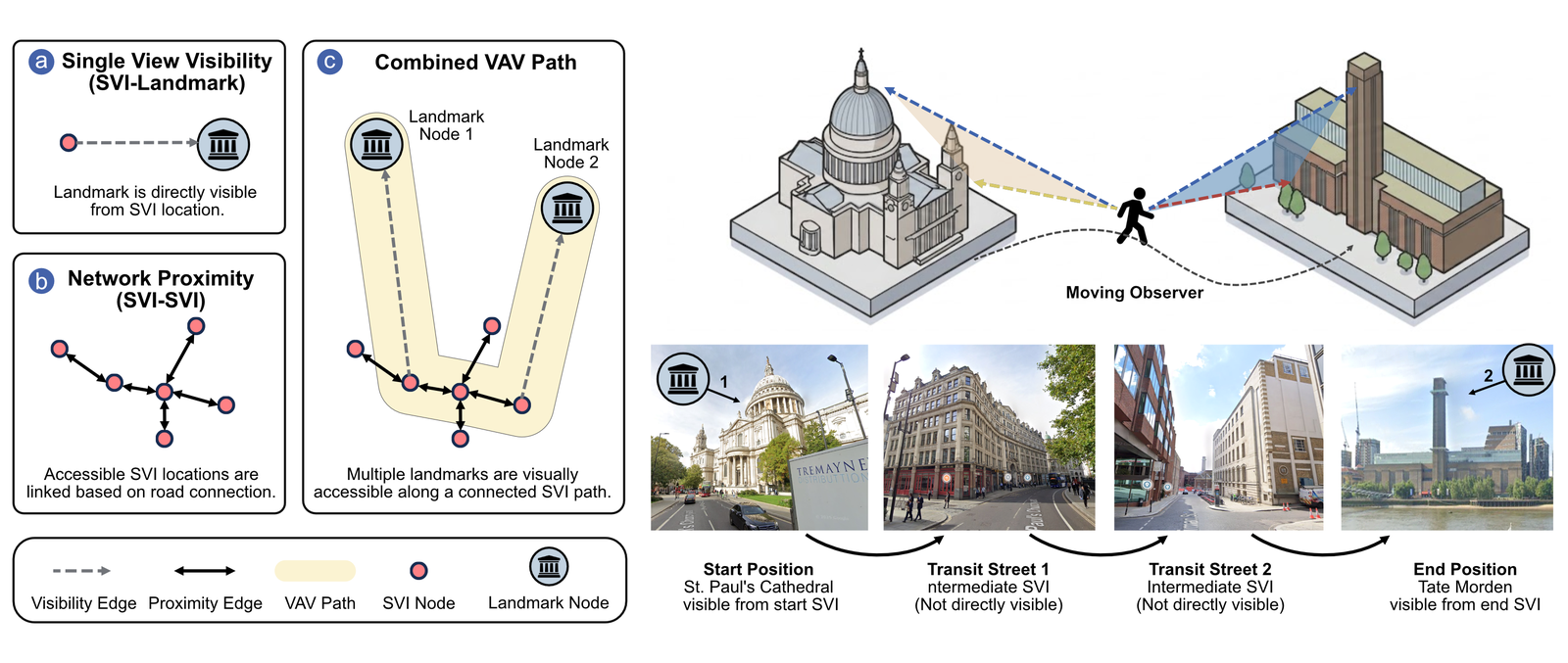}
    \end{subfigure}
    \caption{Left: Different edge types included in the graph definition. Right: An example of VAV path presenting how twp different landmarks can be visually linked with street scenarios. Imagery: Google Street View. }
    \label{fig:edge_types}
\end{figure}

We represent the detected visibility results together with the street-view network in a heterogeneous graph. In this graph, urban landmarks and SVI locations are treated as distinct node types. As shown in the Figure \ref{fig:edge_types}, there are two different types of edges defined in the graph. For adjacent SVI nodes along the road network, undirected edges can be defined to represent their spatial proximity relationships. \ref{sec:sample:appendix_A} elaborates on how to recognise effective proximity connections among individual SVI locations based on road network typology. From SVI nodes to landmark nodes, directed edges represent detected visibility. The graph is constructed using the Python package Deep Graph Library (DGL) \citep{wang_deep_2020}.

With this graph representation, we propose the \textbf{Visible-Accessible-Visible (VAV)} path as the main analytical construct for quantifying spatial-visual links between landmarks. Specifically, we assume that landmarks can be recognised from only a limited number of street-level locations, but these landmark-observation points may frequently appear along different paths. 
Through the constructed graph, and particularly through the defined spatial proximity between street locations and visibility relationships between street locations and landmarks, we can extend landmark visibility from isolated observation points to paths with landmark visibility. On this basis, if a path passes through two locations from which different landmarks are visible, people may build their visual impressions of the landmarks through such pathways. The disruption of some key pathways, even if they are not located at protected landmark viewpoints, may therefore undermine the visual connections between landmarks. Compared with traditional Visibility Graph Analysis (VGA), in which nodes typically represent spatial units and edges represent mutual visibility for investigating the spatial configuration of convex spaces within defined boundaries \citep{turner_isovists_2001}, the VAV paths derived from our graph setting place greater emphasis on semantic visual connectivity among urban elements, as well as how and where these meaningful connections emerge within the street network.

Two extreme cases further illustrate this construct. When such paths are densely distributed, they can represent an areal condition, such as a street block. If paths connecting the visibility of different landmarks are frequently distributed within specific urban blocks, access to these blocks is likely to involve visual encounters with the corresponding landmarks. Conversely, when the path length is reduced to a single point, the construct represents a co-visibility condition, in which multiple urban landmarks can be recognised from the same location. Thus, the VAV path can serve as a flexible framework for simulating visual interactions among multiple landmarks withinn the complex urban road network.

\section{Case Study}

\subsection{Visibility Analysis for Single Landmarks}

\begin{table}[h]
    \centering
    \scriptsize
    \caption{Iconic tall landmark structures investigated in the first case study, including both high‑rise buildings and non‑habitable towers. Imagery: Wikimedia Commons.}
    \label{tab:global_landmark_list}
    \renewcommand{\arraystretch}{1.3}  %
    \begin{tabular}{ccccccc}
        \toprule
         Landmark & The Shard & Petronas Towers & Tokyo Tower & Eiffel Tower & Taipei 101 & Burj Khalifa \\ 
         \midrule
         Query Image &
         \includegraphics[height=3cm]{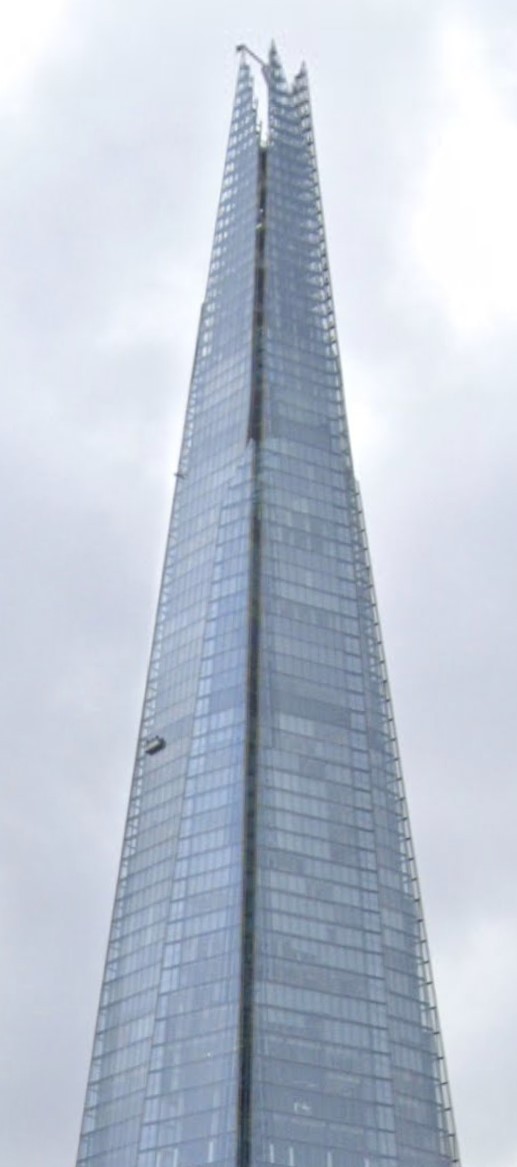} &
         \includegraphics[height=3cm]{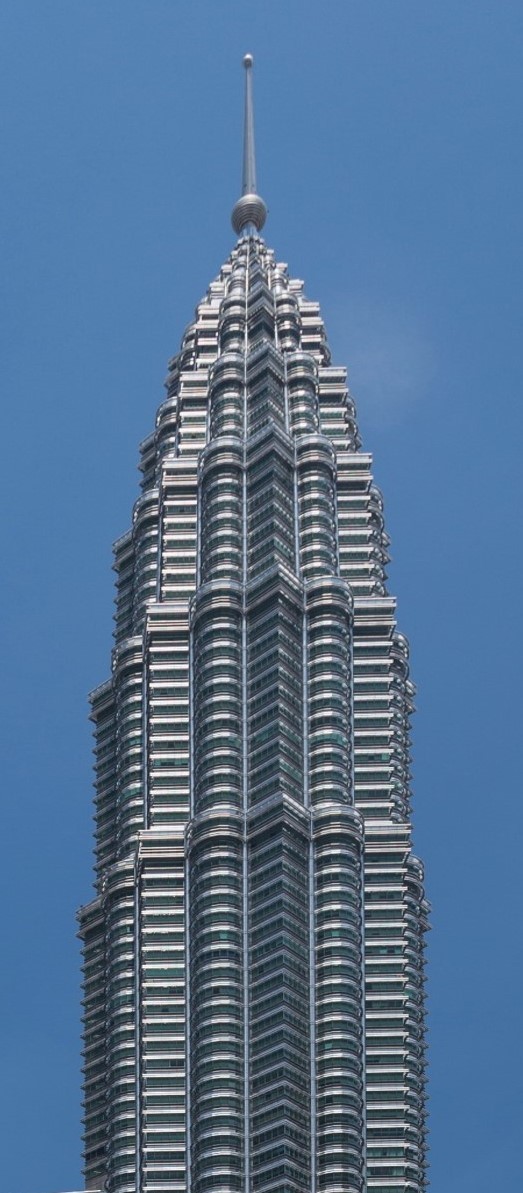} &
         \includegraphics[height=3cm]{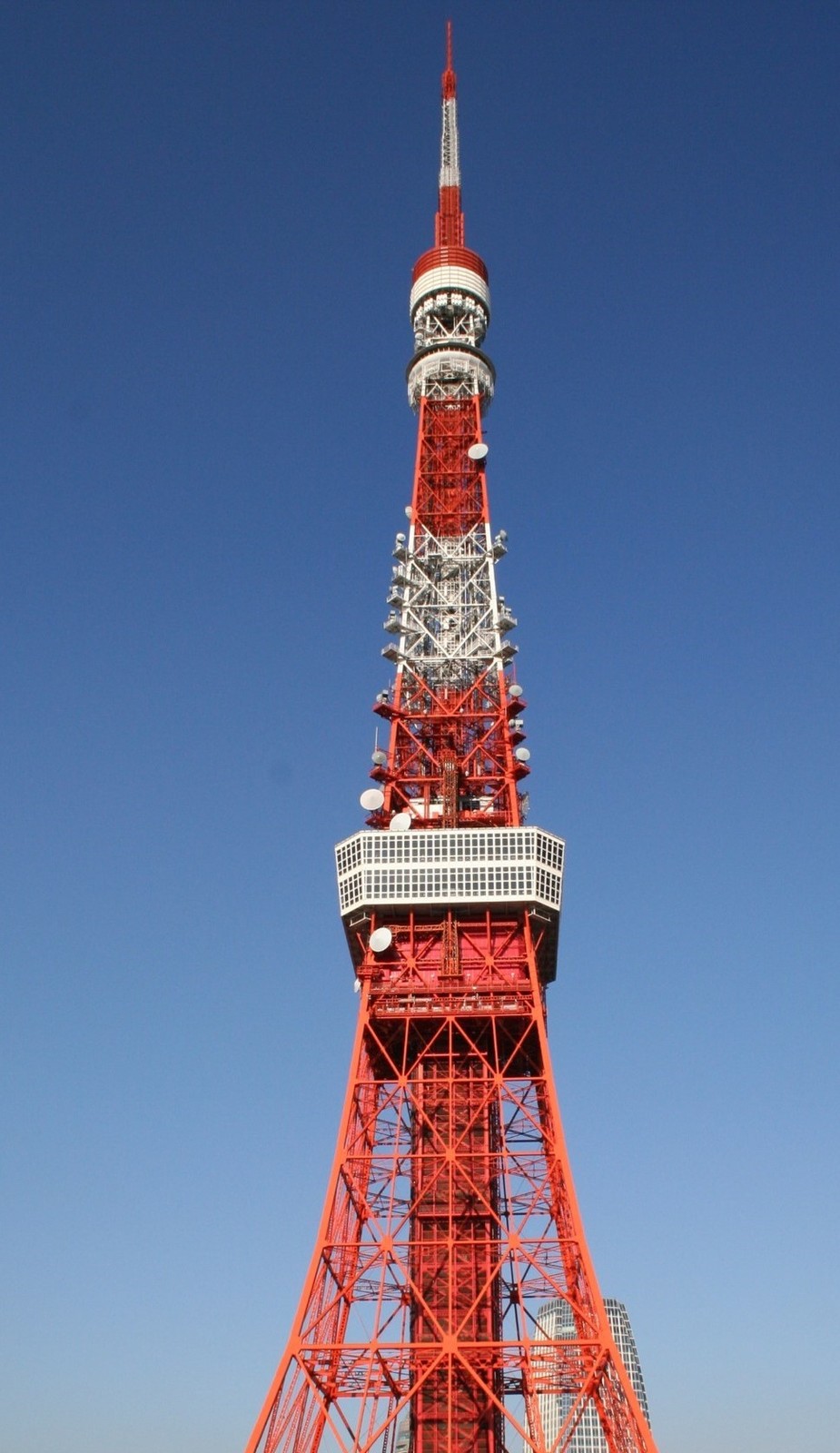} &
         \includegraphics[height=3cm]{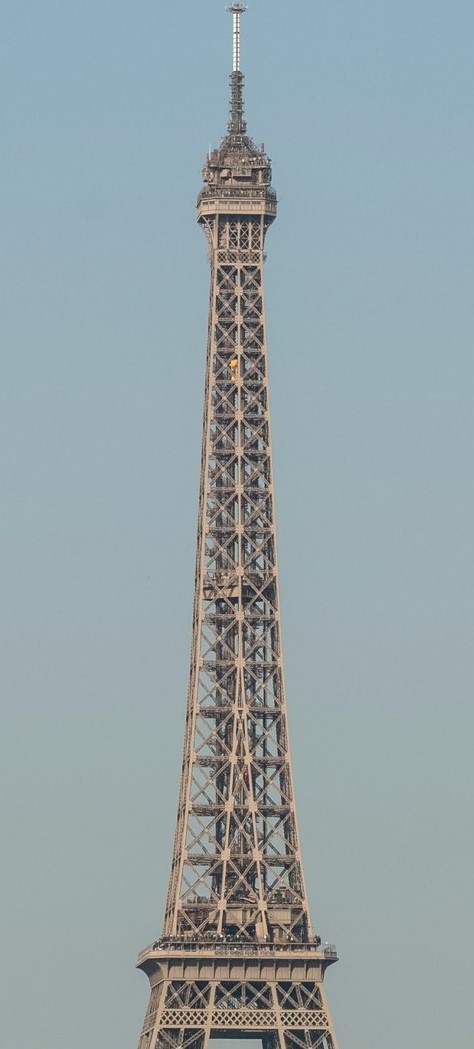} &
         \includegraphics[height=3cm]{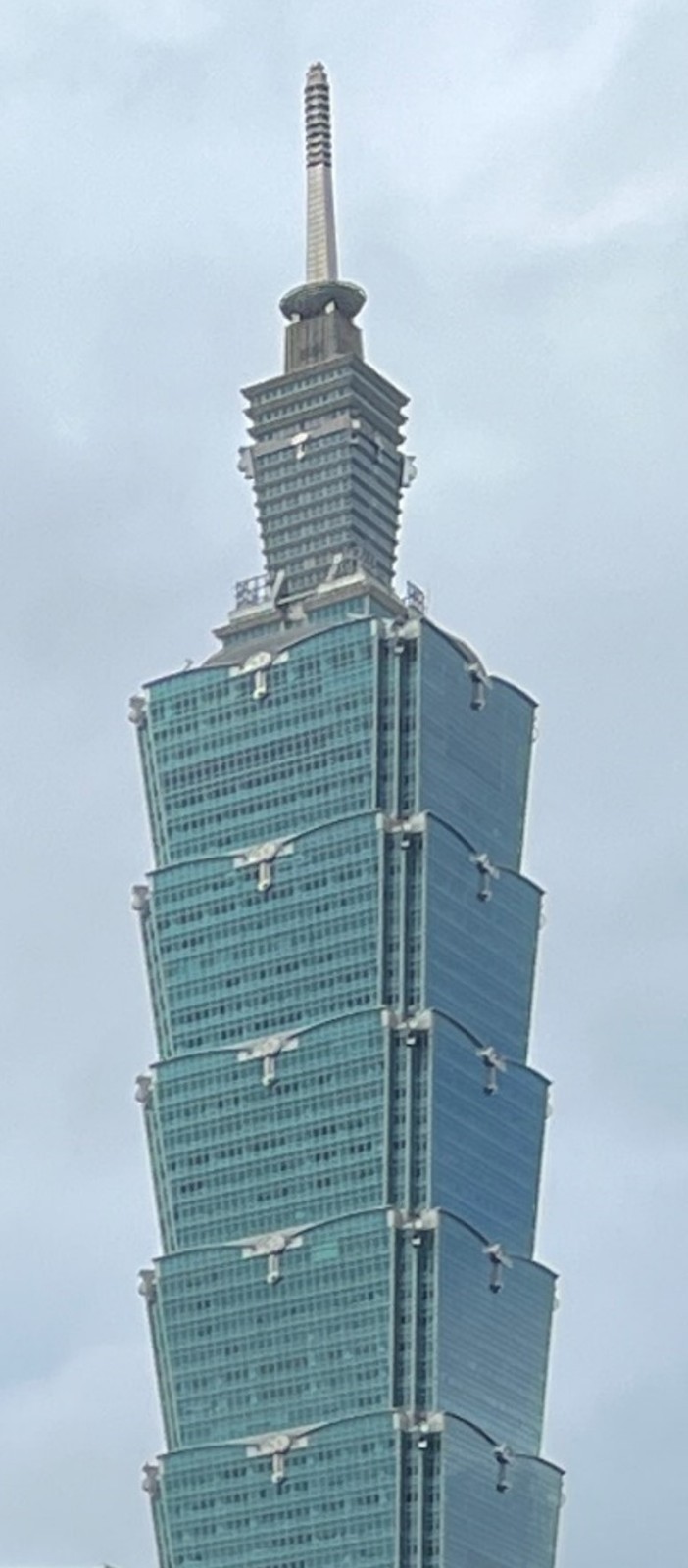} &
         \includegraphics[height=3cm]{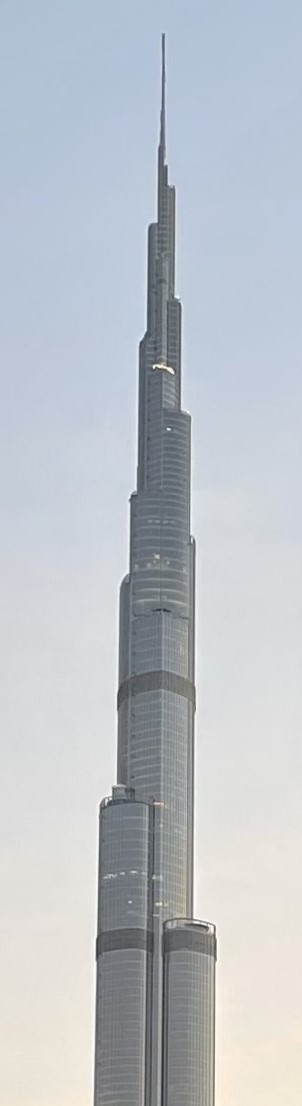} \\ 
         \midrule
         City & London & Kuala Lumpur & Tokyo & Paris & Taipei & Dubai \\
         Height (m) & 310 & 451.9 & 332.9 & 312 & 508 & 829.8 \\
         Built-up Year & 2013 & 1998 & 1958 & 1889 & 2004 & 2010 \\
         Latitude & 51.5045 & 3.1579 & 35.6586 & 48.8584 & 25.0339 & 25.1972 \\
         Longitude & -0.0865 & 101.7113 & 139.7454 & 2.2945 & 121.5645 & 55.2744 \\
         \bottomrule
    \end{tabular}
\end{table}

In the first case study, we assess the effectiveness of the image-based visibility detection method by applying it to six well-known tall landmark structures in six global cities. The details of these landmarks are presented in Table~\ref{tab:global_landmark_list}. For each landmark, sampling points are generated at 30 m intervals within a 3000 m buffer based on OSM road-network data, and their coordinates are used to retrieve the nearest SVI from the Google Street View service. For London, Tokyo, and Paris, more than 84.1\% of the images were retrieved between April and October, reflecting a balance between reducing seasonal bias and maintaining sufficient spatial coverage, as stricter seasonal constraints would limit the availability of usable SVI. Taipei, Kuala Lumpur, and Dubai are expected to be less affected by seasonal bias due to their subtropical, tropical, and desert climates, respectively. The spatial and temporal distribution of collected SVIs is detailed in Figure~\ref {svi_metadata}. Steps 1--4 introduced in Section~\ref{sec:image_based_visibility} are then applied to detect landmark visibility from the corresponding SVI. To evaluate the detection performance, we use the manually labelled validation dataset described in Section~\ref{sec:sample:appendix_B}. To avoid a circular argument, the validation dataset is completely independent from any model inference results. Precision, recall, F1-score, and balanced accuracy are computed to evaluate the method's performance. In addition, ablation experiments are conducted to validate the effectiveness of the zoom-in image preprocessing strategy and the use of image queries in the landmark visibility detection workflow.

We compare the SVI-based visibility with visibility simulated from open 3D urban models and, as an additional reference, with the distribution of geotagged Flickr images. Firstly, we compare the image-based visibility results with the 3D simulated visibility results at SVI locations, by evaluating the classification accuracy metrics for `visible' or `invisible' with the same validation dataset created. We carry out 3D visibility simulation with the Python package of `VoxCity'\footnote{\url{https://github.com/kunifujiwara/VoxCity}} \citep{fujiwara_voxcity_2026}. `VoxCity' is a one-stop toolset for collecting and converting open 3D urban data to voxels, and conducting 3D spatial analysis based on them. 
3D visibility simulation is carried out within a 3000-meter radius from each landmark, using a uniform 3D voxel grid size of 5 × 5 × 5 m. The data models applied to generate voxels incorporate buildings, tree canopies and terrain, whose sources are detailed in Table~\ref{tab:voxcity_source_vertical} in Appendix \ref{sec:sample:appendix_C}. We used different data sources for each city, depending on the sources' geographic coverage and data quality. 

Beyond the direct comparison of visibility labels, we further examine the potential reasons for discrepancies between the SVI-based and 3D model-based methods. We examine the proportions of key street view elements, such as buildings, greenery, and sky, at locations classified as ‘only SVI visible’, ‘only 3D visible’, ‘both visible’, and ‘both invisible’. These semantic elements are extracted from SVI using the Python package `ZenSVI'\footnote{\url{https://github.com/koito19960406/ZenSVI}}, developed by \citet{ito_zensvi_2025}. This analysis helps reveal how differences in foreground obstruction, visual openness, and scene composition may contribute to the discrepancies between image-based and 3D-based visibility results.

The above comparison should be interpreted within two important scope conditions. First, due to the limited availability and inconsistent coverage of high-quality 3D urban models across cities, `VoxCity' is used here as a representative open-data-based 3D workflow for comparison, rather than as a benchmark for all possible 3D visibility simulation methods. Second, because no citywide ground-truth visibility dataset is available that can be equally applied to both image-based and 3D-based approaches, the quantitative comparison is restricted to SVI locations where image observations are available. %
This design may not fully capture cases where 3D simulation could estimate visibility in areas not covered by SVI, but it provides a consistent basis for evaluating how the proposed image-based method performs under real street-level observation conditions.

As an additional reference, we collect geotagged images with landmark tags from Flickr, to recognise their distribution difference with landmark visibility derived from SVI and the open 3D models. Crowd-sourced images from Flickr have been proven to have a strong association with public visiting towards urban space \citep{wood_using_2013,mor_3d_2021} and landscape preference \citep{foltete_coupling_2020}. The Flickr images with landmark tags are searched with Flickr's official API \footnote{\url{https://www.flickr.com/services/api/}}, and within a 3000 m radius of the landmark.

\subsection{Multi-landmark Visibility and the Visual Connection}

\begin{figure}[htbp]
    \centering
    \begin{subfigure}{1\textwidth}
        \centering
        \includegraphics[width=\textwidth]{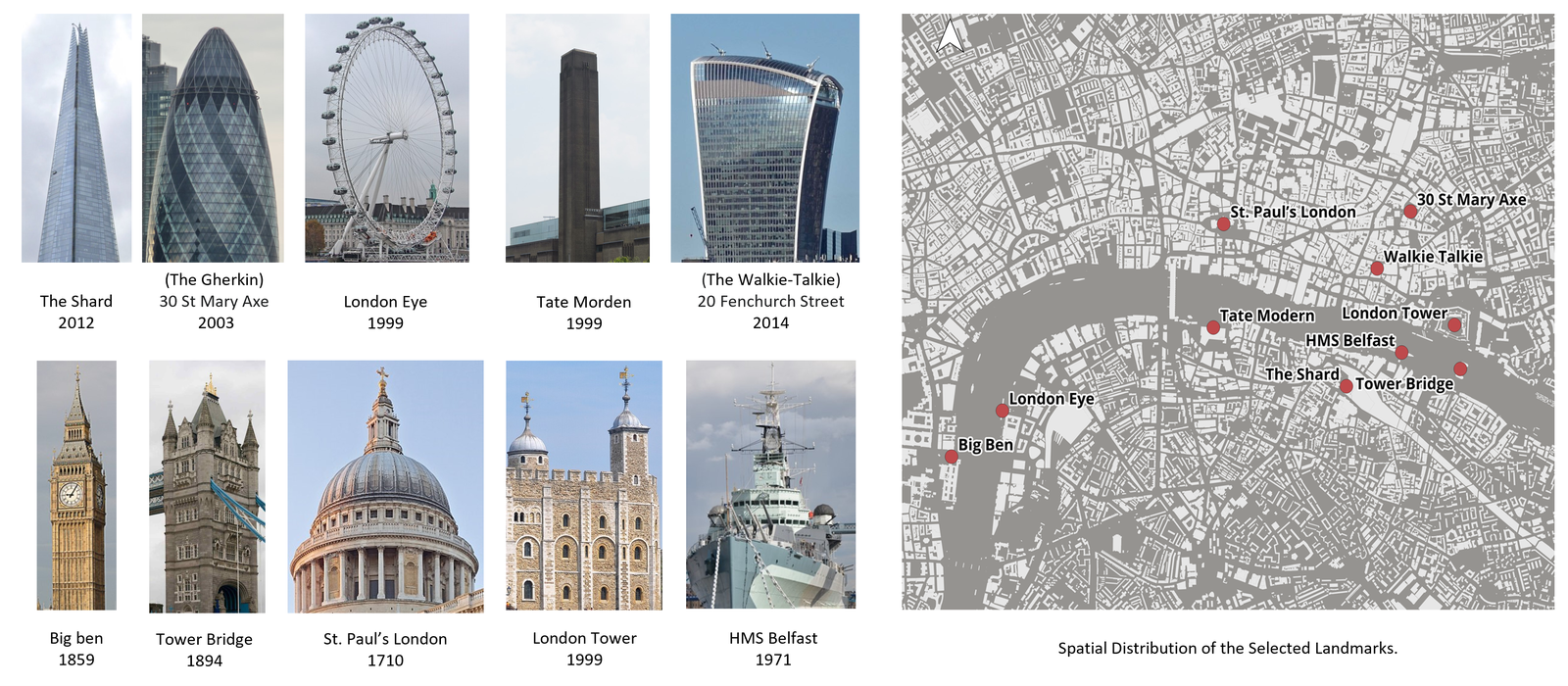}
    \end{subfigure}
    \caption{Left: Query images of different landmarks investigated in the second case study. Imagery: Wikimedia Commons. Right: The spatial distribution of the selected landmarks along the River Thames. Basemap: © OpenStreetMap contributors. }
    \label{fig:london_landmarks}
\end{figure}

In the second case, we explore our method's potential in revealing complex visual structure and connections formed by multiple landmarks, and its value in heritage conservation.
With the latest SVI collected from Google Street View, we investigate the visibility of 10 famous landmarks along the River Thames in London, UK, and aim to reveal their spatial-visual interaction with each other and with the street space. The landmarks are selected with reference to the review count ranking on Tripadvisor. The details and distribution of the landmarks are illustrated in Figure \ref{fig:london_landmarks}.
After detecting visibility for each landmark with the proposed image-based method, we follow steps in Section \ref{sec:graph_construction} to build a heterogeneous graph. 
We conduct graph-based analyses to understand the direct and indirect visual connections among landmarks. %

We apply a random-walk-based approach \citep{pearson_problem_1905,xia_random_2020} to identify VAV paths linking different landmarks. Starting from a street-view location that is visually connected to a given landmark, the algorithm stochastically explores the road network for a predefined number of steps and records a successful transition when the walk reaches a location from which another landmark becomes visible. To avoid unrealistic movements, an angle-penalty strategy is incorporated into the path search so that sharp reversals and U-turns are discouraged.

By normalising the number of valid walks between each ordered pair of landmarks by the total number of simulated walks, we derive a directional measure of visual connection strength that is comparable across landmark pairs. The resulting connection matrix captures both asymmetry and hierarchy in inter-landmark visibility: some landmarks more frequently act as visual origins or viewing platforms, while others more often emerge as visual destinations. This is particularly relevant for the selected landmarks along the Thames, which function simultaneously as components of the urban skyline and as important observation points. Furthermore, the spatial distribution of successful VAV paths reveals corridors that are repeatedly involved in linking landmark pairs. For local planners, such evidence is critical for maintaining existing landmark visibility patterns and cross-landmark visual relationships.

To evaluate parameter robustness, we test a grid of walk lengths (20, 40, 80, 120, and 160 steps) and numbers of walks (500, 1,000, 1,500, and 2,000), each repeated with five random seeds. For each parameter combination, we construct the landmark-to-landmark connectivity matrix and assess its stability across seeds using three complementary metrics. First, matrix correlation, computed using the Pearson correlation coefficient, measures whether different runs preserve a similar overall pattern of pairwise landmark relations. Second, the relative Frobenius norm quantifies the magnitude of matrix-level differences, and is therefore sensitive to changes in the absolute strengths of connections. Third, top-$k$ edge Jaccard similarity evaluates the reproducibility of the strongest directional relations by comparing the overlap among the highest-ranked landmark pairs across runs. Together, these metrics assess the global consistency of the connectivity structure and the robustness of the most salient inter-landmark links. The angle-penalty strategy and the metric calculations are detailed in \ref{sec:random_walk}. The relationship matrix is visualised using the mean connectivity matrix obtained with a walk length of 80 and 2,000 walks, which proved robust from the parameter evaluation. The spatial distribution of successful VAV paths is visualised under the same parameter setting.

\section{Results}

\subsection{Effectiveness of Image-based Visibility Analysis}
\subsubsection{Spatial and Distance Distributions}
\label{sec:single_landmark_visibility}

\begin{figure}[htbp]
    \centering

    \begin{subfigure}{\textwidth}
        \centering
        \makebox[\textwidth][c]{%
            \includegraphics[width=1.2\textwidth]{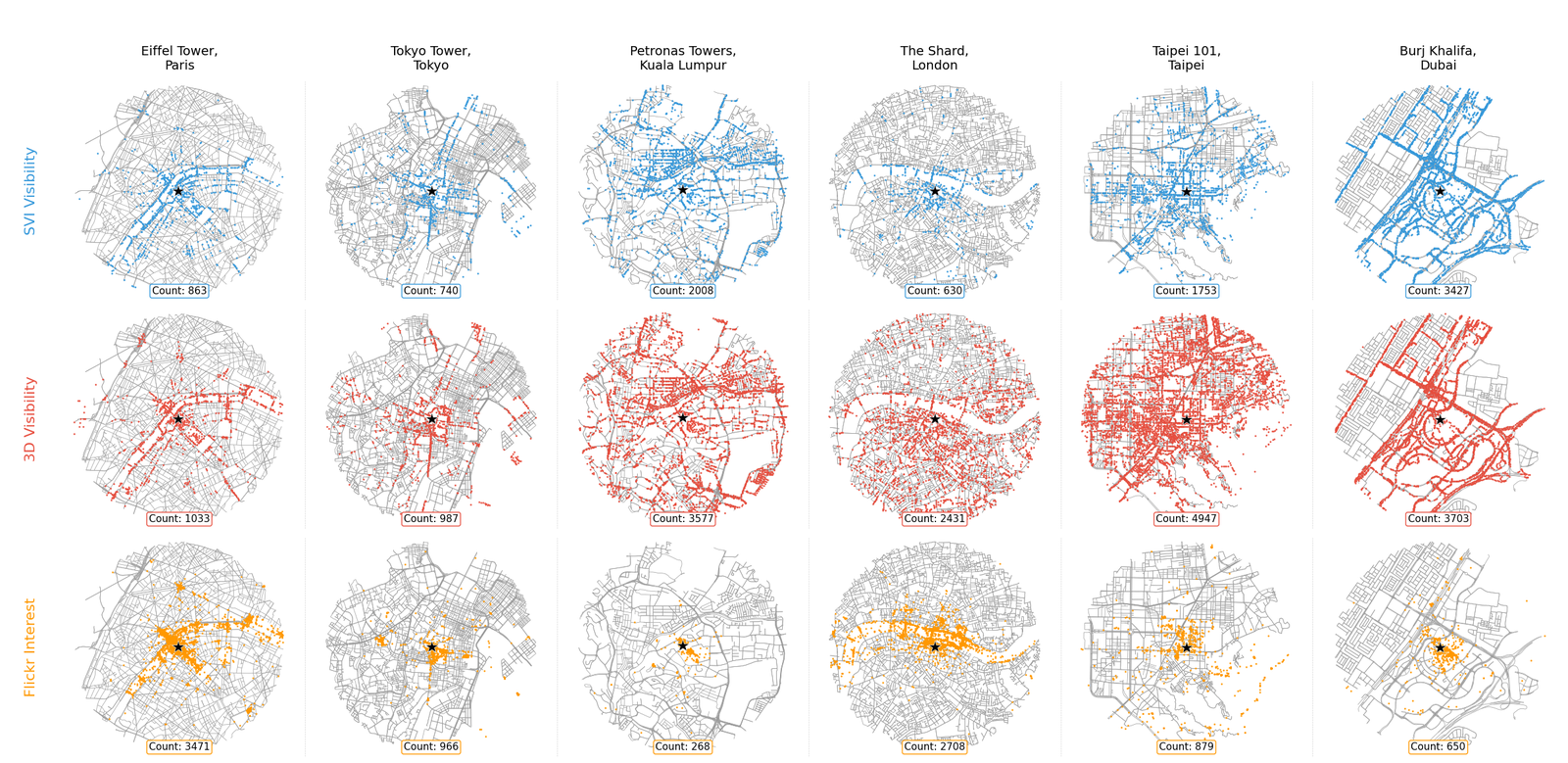}
        }
        \caption{A comparison of distribution for SVI-based landmark visibility, 3D simulated landmark visibility, and Flickr images with landmark tags. Basemap: © OpenStreetMap contributors.}
        \label{fig:3_visibility_comparison}
    \end{subfigure}

    \vspace{0em}

    \begin{subfigure}{\textwidth}
        \centering
        \makebox[\textwidth][c]{%
            \includegraphics[width=1\textwidth]{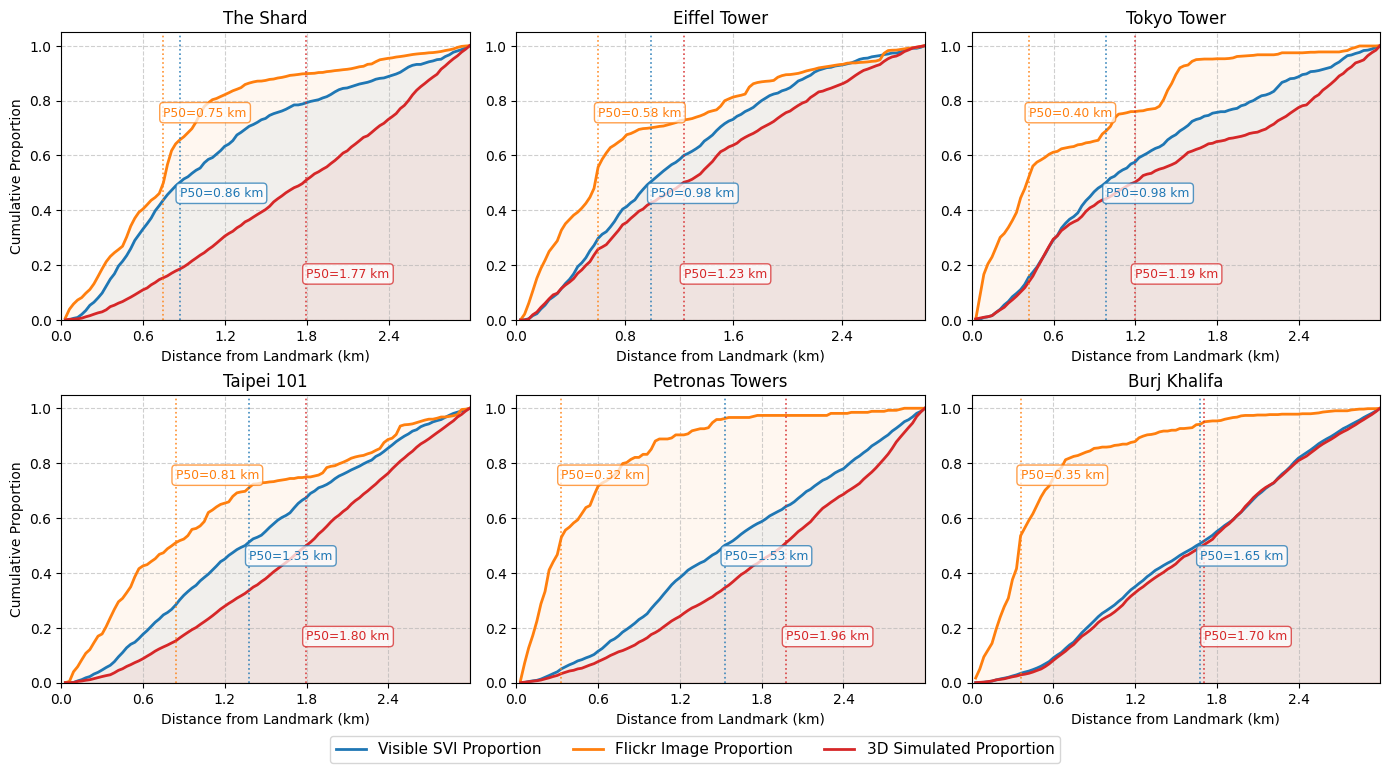}
        }
        \caption{Accumulated proportion distribution of landmark-visible SVI locations and Flickr images along the distance from the corresponding landmarks. The distances where the cumulative proportions reach 50\% are labelled.}
        \label{fig:visibility_along_distance}
    \end{subfigure}

    \caption{Comparison of landmark visibility distribution and cumulative proportion distribution along distance from landmarks.}
    \label{fig:combined_visibility}
\end{figure}

Figure \ref{fig:3_visibility_comparison} presents the distributions of SVI locations detected as visible to the corresponding landmarks. The distributions are compared with two other visibility proxies: visibility simulated with 3D models and the distributions of Flickr images with the landmark tags. For 3D simulated visibility, it is represented based on $5 \times 5$ m grids. Grids identified as visible are joined spatially to the SVI locations for comparison. The initial analysis results for SVI-based visibility and 3D-simulated visibility are presented in Figure \ref{fig:raw_visibility_map} in Appendix. 

Overall, landmark-based visibility and 3D simulated visibility share more similarities in spatial distribution. In particular, the visible-location patterns in Paris, Tokyo, and Dubai are similar in both extent and density. In Kuala Lumpur, London, and Taipei, the open 3D workflow identifies more visible locations than the SVI-based approach. In contrast, the Flickr distributions are generally more compact and more strongly concentrated around the landmarks.

Figure \ref{fig:visibility_along_distance} compares the cumulative proportion distributions of landmark-visible SVI locations, 3D-simulated visible locations, and Flickr images over distance from the corresponding landmarks. Across the cases, the SVI-based curves consistently fall between the Flickr and 3D curves. Similar to Flickr, the SVI-based curves show a clear distance-decay pattern; however, this decay is less concentrated near the landmark. Similar to the 3D results, the SVI-based curves indicate a broader spatial spread of visibility, but they do not approach a uniform distribution over distance. In this sense, SVI-based visibility occupies an intermediate position between highly concentrated Flickr image-taking and more spatially extensive 3D-simulated visibility.

Another insight from the comparison is that image-taking behaviour, reflected by the spatial distribution of Flickr images, does not necessarily coincide with the broader distribution of landmark visibility reflected by the other two methods. For The Shard, Eiffel Tower, Tokyo Tower, and Taipei 101, all three visibility proxies decline rapidly with distance and follow broadly similar patterns. Although the Flickr curves generally accumulate faster, this overall alignment suggests that the distance ranges from which these landmarks are visible at street level are also broadly the ranges where people tend to photograph them. Among the four cases, The Shard shows the closest alignment between SVI-based visibility and Flickr distribution, with the smallest difference in the distances at which the cumulative proportion reaches 50\%. By contrast, the Petronas Towers and Burj Khalifa show a different pattern. Flickr images tagged with these landmarks are concentrated much more strongly near the buildings, mostly within 500 m, whereas SVI-based visibility remains distributed across a wider distance range, with many visible locations beyond 1000 m. This contrast suggests that, for some landmarks, image-taking is more destination-oriented, while street-level visibility remains widespread across everyday urban locations.

\subsubsection{Performance Evaluation and the Impact of Occlusions}

\begin{table}[h]
    \scriptsize
    \centering
    \caption{A summary of performance for image-based visibility detection and 3D model-based visibility detection.}
    \begin{tabular}{lccccccc}
        \toprule
        & & \multicolumn{3}{c}{Image-based Visibility Detection} & \multicolumn{3}{c}{3D Model-based Visibility Simulation} \\
        \cmidrule(lr){3-5} \cmidrule(lr){6-8}
        Class & Support & Precision & Recall & F1-score & Precision & Recall & F1-score \\
        \midrule
        Invisible & 1807 & 0.9652 & 0.8600& 0.9096& 0.929 & 0.7023 & 0.7999 \\
        Visible & 593 & 0.6797 & 0.9056 & 0.7766 & 0.4797 & 0.8364 & 0.6097 \\
        \midrule
        Accuracy &  & \multicolumn{1}{c}{} & 0.87 & \multicolumn{1}{c}{} & \multicolumn{1}{c}{} & 0.74 & \multicolumn{1}{c}{} \\
        \bottomrule
    \end{tabular}
    \label{tab:visibility_comparison}
\end{table}

To validate the effectiveness of the proposed image-based visibility detection method, we first conduct ablation experiments on the validation dataset to investigate two key design choices: (1) the impact of the zoom-in crop strategy, and (2) the difference between image queries and text queries for zero-shot landmark detection. The results are reported in Table \ref{tab:visibility_ablation} and in Figure \ref{fig:ablation_six_landmarks}. The results show that the zoom-in preprocessing substantially improves the performance of the zero-shot object detection model. In addition, image queries generally outperform text queries in precision, recall, and F1-score for both visible and invisible classes, indicating their advantage in detecting landmarks from complex street view scenes.

We further compare the performance of image-based visibility detection and 3D model-based visibility simulation on the validation dataset. The result is summarised as in Table~\ref{tab:visibility_comparison}. Under this validation setting, the image-based method achieves higher overall accuracy and higher visible-class precision than the open 3D workflow, while both methods perform more strongly on the invisible class than on the visible class.

\begin{figure}[htbp]
    \centering

    \begin{subfigure}{1\textwidth}
        \centering
        \includegraphics[width=0.95\textwidth]{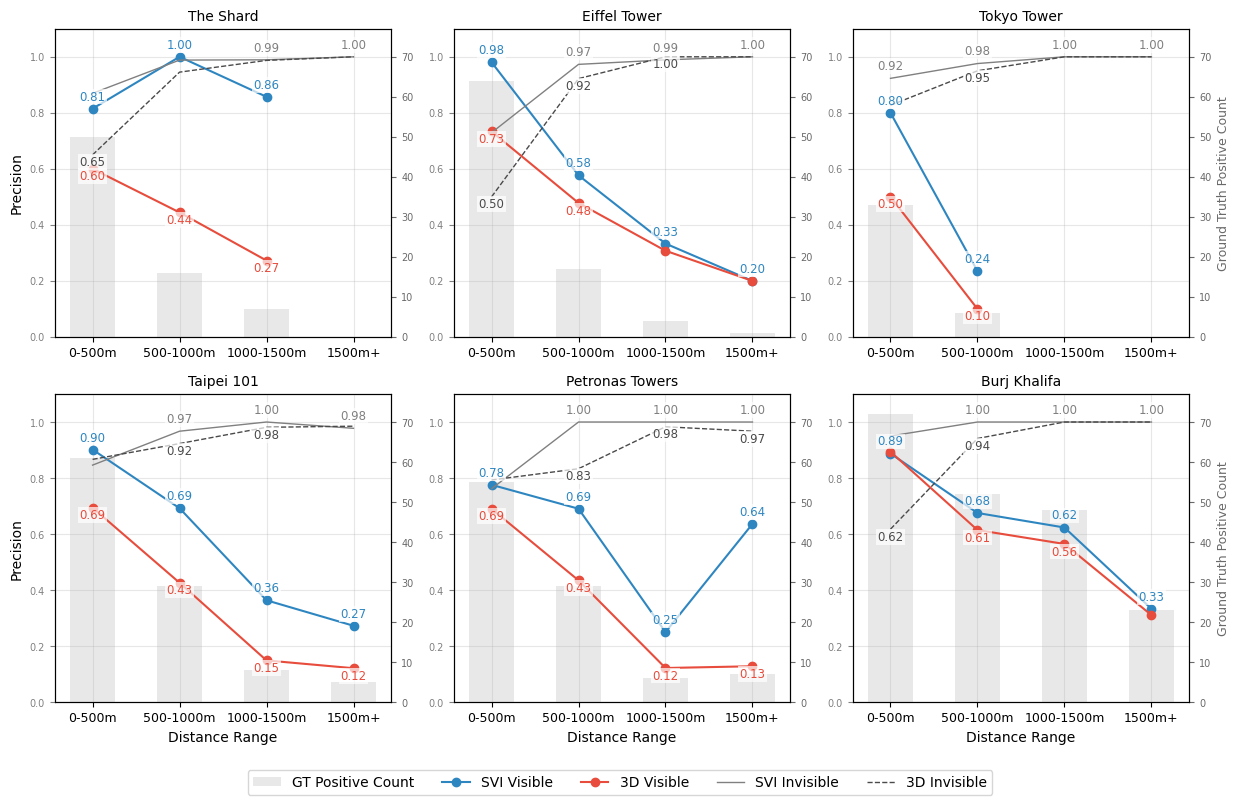}
        \caption{Line plots comparing the precision of landmark visibility detection based on image and 3D methods.}
        \label{fig:line_plots_precision}
    \end{subfigure}

    \begin{subfigure}{\textwidth}
        \centering
        \includegraphics[width=0.95\textwidth]{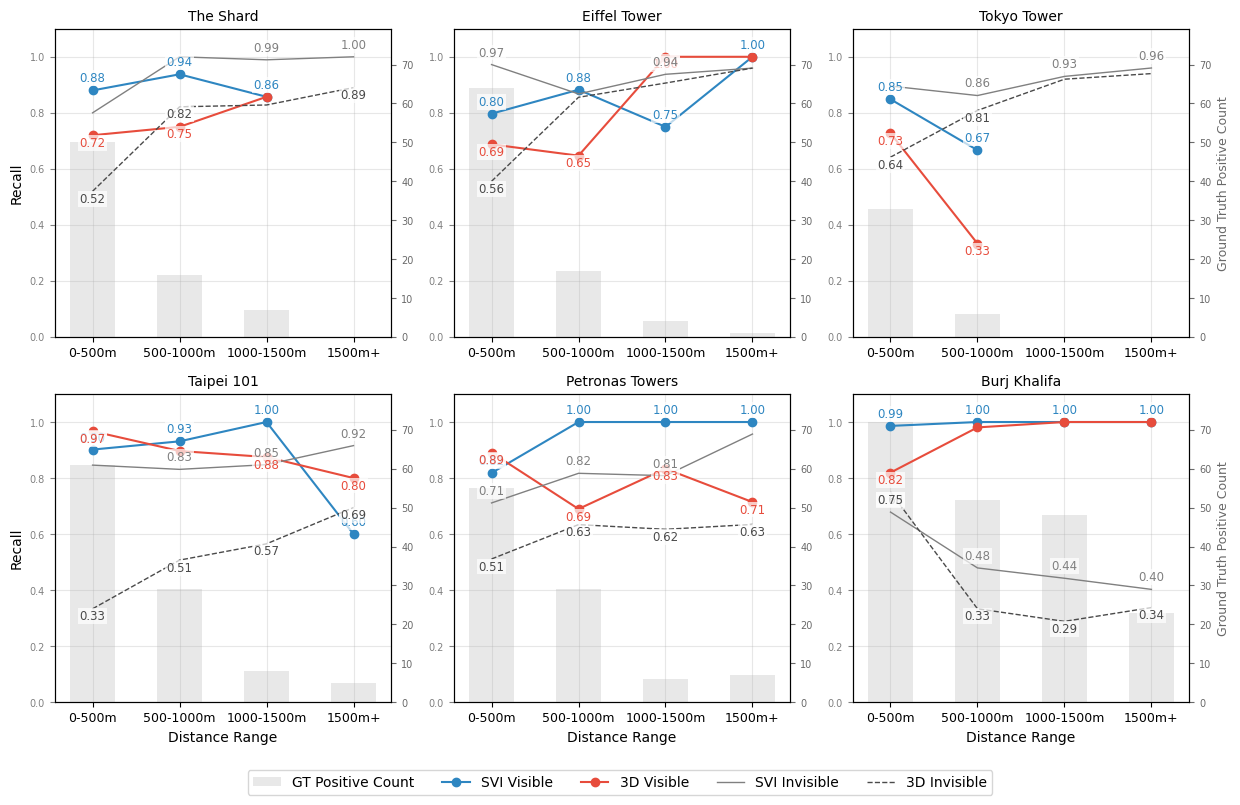}
        \caption{Line plots comparing the recall of landmark visibility detection based on image and 3D methods.}
        \label{fig:line_plots_recall}
    \end{subfigure}

    \caption{Plots comparing the performance in SVI-based and 3D-based visibility detection, across different landmarks and distance bands. For each landmark, the validation set is created by randomly sampling 100 SVIs from each distance band.}
    \label{fig:svi_vs_3d}
\end{figure}

To gain a more detailed view of both methods, Figure~\ref{fig:line_plots_precision} and Figure~\ref{fig:line_plots_recall} report the class-specific precision and recall of the SVI-based and 3D-based methods across different landmarks and distance bands. The grey bars indicate the number of ground-truth visible samples in each distance band, providing additional context for interpreting the performance changes.
Compared with the 3D-based method, the SVI-based method generally achieves higher or comparable visible-class precision across landmarks and distance bands.
For most landmarks, the SVI-based method achieves its strongest performance in detecting visible locations within the 0--500 m distance band. In this range, both precision and recall are generally around or above 0.8. The Eiffel Tower presents the highest visible-class precision, reaching 0.98. As distance increases, the precision of visible detection decreases for several landmarks. For example, within the 500--1000 m distance band, the precision drops to around 0.6 for the Eiffel Tower, Burj Khalifa, Petronas Towers, and Taipei 101, and further drops to 0.24 for Tokyo Tower. The Shard is an exception, where the detection performance remains relatively stable across distance bands.
This decline in precision with increasing distance is largely associated with the decreasing proportion of truly visible samples in the validation dataset. As visible samples become less frequent at longer distances, a small number of false-positive predictions can have a larger effect on the precision score.

The recall results provide a complementary perspective. The recall of the SVI-based method is generally less sensitive to increasing distance. For several landmarks, the method maintains high recall for visible locations even in farther distance bands, suggesting that it remains effective in retrieving visible landmarks once they appear in the image. The recall for invisible locations is also generally stable across distance bands. Nevertheless, distance still introduces uncertainty, especially when the landmark occupies only a small portion of the image or becomes visually ambiguous. For example, for the Burj Khalifa, both methods are prone to overestimating the visible locations at increasing distances.
To further reduce the influence of class imbalance, we also summarize balanced accuracy across different landmarks and distance bands, as shown in Figure~\ref{fig:additional_line_plots}. This additional metric again indicates that the SVI-based method provides effective and relatively more robust performance than the 3D-based method at the shared SVI observation locations.

\begin{figure}[htbp]
    \centering
    \begin{subfigure}{\textwidth}
        \centering
        \includegraphics[width=\textwidth]{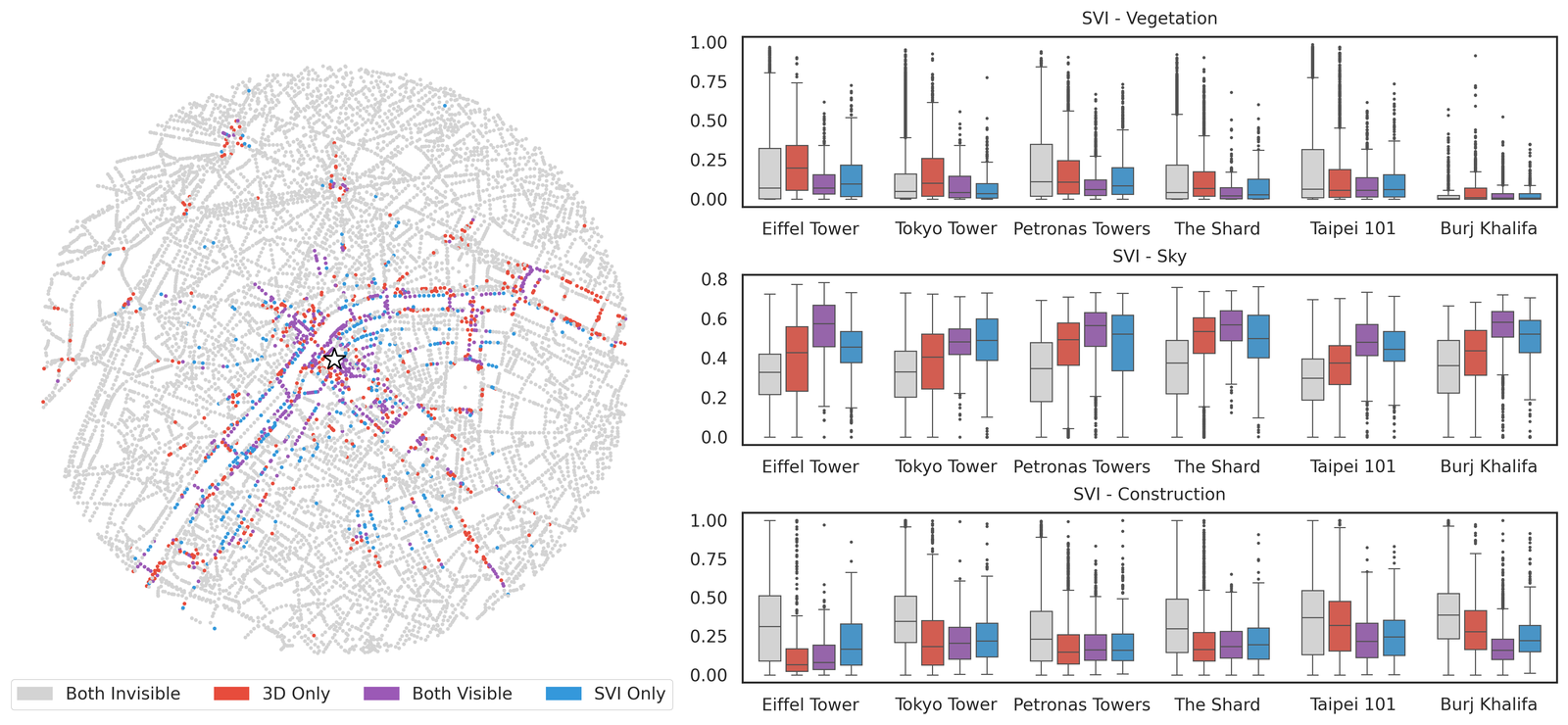}
    \end{subfigure}
    \caption{Left: Spatial distribution of match and mismatch locations between SVI-based visibility and 3D-based visibility for the Eiffel Tower.
    Right: Box-plots illustrating the proportion of semantic elements in the foreground of the views.}
    \label{fig:combined_boxplot}
\end{figure}

To further explain the difference between SVI-based and 3D-based visibility detections, Figure \ref{fig:combined_boxplot} illustrates how the visibility detection outcomes relate to the visible semantic elements of vegetation, sky, and construction in the foreground. 
SVI locations are reclassified into `Both Invisible', `Both Visible', `3D Only', and `SVI Only'.
Grouped with the classification, box-plots on the right side illustrate the distribution difference of semantic elements of vegetation, sky and construction, across different landmarks. As a showcase, on the left side, the spatial distribution of the classifications is mapped for the Eiffel Tower.
Both methods consistently demonstrate that areas where landmarks are visible tend to be open spaces. These areas are characterised by a significantly higher visible sky ratio and lower construction and vegetation ratios. In contrast, regions where landmarks are not visible typically present a more confined and crowded visual scene, with comparatively higher visible construction ratios. This overall trend reflects a general urban visual pattern that transcends the differences in the detection methods. Compared to locations detected visible only by the 3D method, those only identified as visible via SVI generally exhibit less vegetation obstruction. This discrepancy is particularly pronounced in observations toward landmarks such as the Eiffel Tower, Tokyo Tower, and The Shard. 
This finding suggests that the SVI-based method more effectively considers tree obstructions, a factor not as well captured by the conventional 3D approach. 
At the same time, the construction ratios at SVI-detected locations do not show a lower trend compared to those from the 3D method. This justifies that the SVI approach is not biased toward low-density building areas; it can handle complex and crowded urban environments with notable built obstructions.

\subsection{Uncovering Visual Connection and Interaction Among Multiple Landmarks}
In Section \ref{sec:single_landmark_visibility}, we demonstrated the reliability of SVI in capturing the visibility of individual landmarks and its unique potential to reveal the visual context. Using 10 famous landmarks along the River Thames in London, in this section, we examine SVI's ability to assess their visual relationships with one another and their interaction with street space driven by visibility.

\subsubsection{Different Roles of Landmarks along the River Thames}

\begin{figure}[htbp]
    \centering
    \makebox[\textwidth][c]{%
        \includegraphics[width=1.2\textwidth]{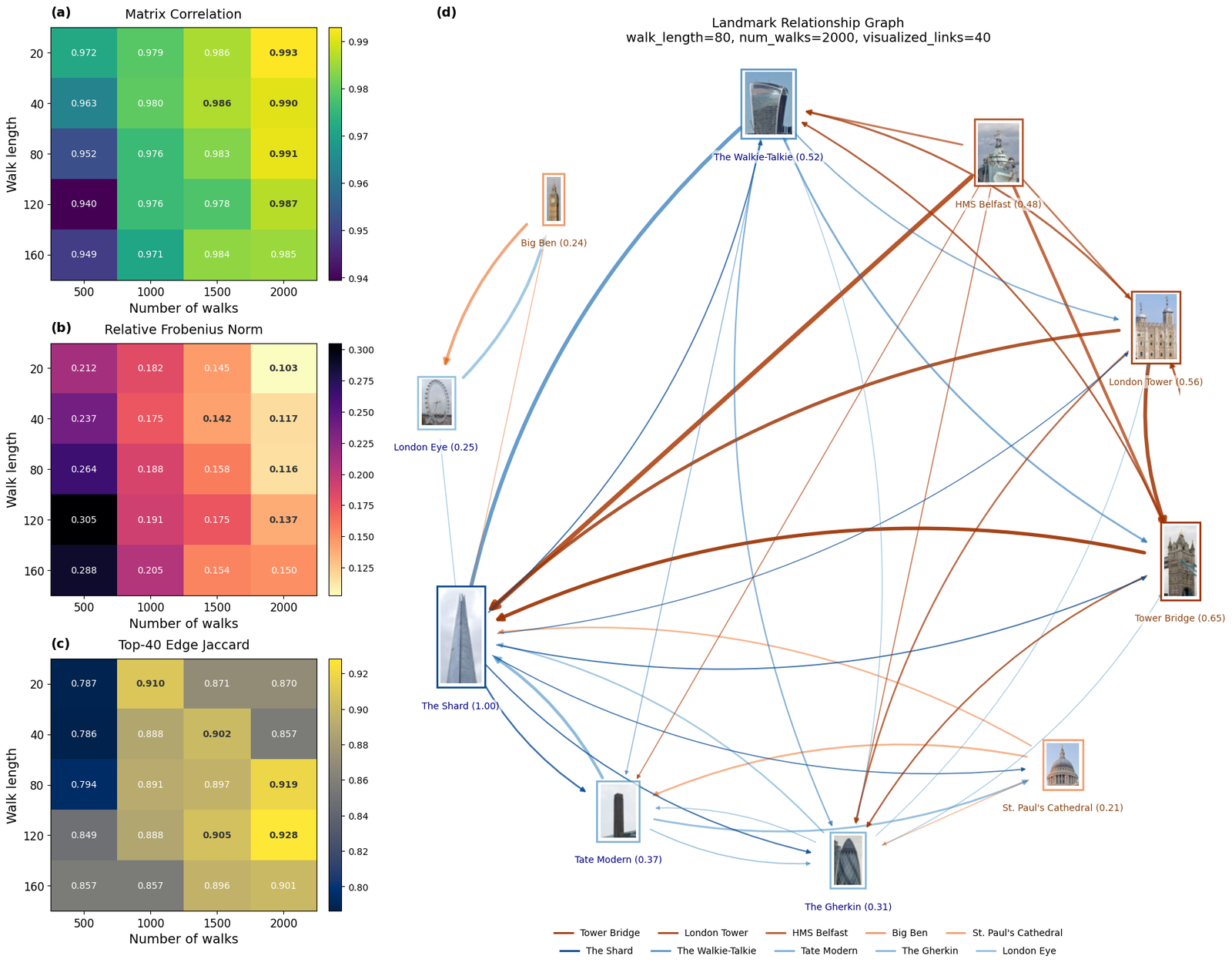}%
    }
    \caption{(a) Matrix correlation of landmark-to-landmark connectivity across different walk lengths and numbers of walks. 
(b) Relative Frobenius norm of connectivity matrices across parameter settings. 
(c) Top-\(k\) edge Jaccard similarity of the strongest directional landmark relations across parameter settings. Here, we pick $k$ equals to 40 to ensure a wide coverage of landmark pairs while preventing overcomplicated analysis and visualization.
(d) Directed visual connections between modern and historical landmarks along the River Thames, visualised from the mean connectivity matrix under the selected parameter setting (\(80\) steps and \(2{,}000\) walks); node size indicates normalized landmark connectivity strength, and edge width and transparency indicate relative directional connection strength. Imagery: Wikimedia Commons. Software: NetworkX \citep{hagberg_exploring_2008}.}
    \label{fig:intervisibility_figures}
\end{figure}

Figure~\ref{fig:intervisibility_figures}(a-c) first evaluates the robustness of the random-walk-based VAV simulation under different parameter settings. 
The three heatmaps summarize the consistency of the landmark-to-landmark connectivity matrices across different walk lengths and walk counts, with different random seeds. 
It is revealed that stability generally improves as the number of walks increases, while excessively short walk lengths tend to produce less consistent structures. Among the tested parameter combinations, the setting of \(80\) steps and \(2{,}000\) walks provides a suitable balance between high structural consistency, relatively low matrix-level variation, and stable recovery of the strongest directional links. We therefore adopt this parameter setting for the subsequent visualization and interpretation of visual connections among landmarks.

Based on the mean connectivity matrix under \(80\) steps and \(2{,}000\) walks, Fig.~\ref{fig:intervisibility_figures}(d) visualizes the directional visual connections between modern and historical landmarks along the River Thames, specifically the likelihood that one landmark can be reached visually from another through the surrounding street network and sequential views. Arrows indicate the direction of visual connection, and their widths represent the relative strength of these path-based relations. The results suggest that The Shard more frequently appears as a visually dominant destination within the urban scene, being more often encountered from other locations than serving as an effective origin of broader landmark visibility. By contrast, the area around The Walkie-Talkie and HMS Belfast functions more strongly as a vantage zone from which multiple other landmarks can be visually connected. London Tower and Tower Bridge emerge as important visual hubs, playing dual roles as both strong observation origins and prominent visual destinations. St~Paul's Cathedral and Tate Modern, as well as Big Ben and the London Eye, exhibit more localized and comparatively weaker links, indicating that their visual relationships are more spatially constrained by street configuration, relative height, and viewing opportunity. Overall, the path-based connectivity results reveal not only which landmarks are visually prominent, but also how different parts of the Thames-side urban landscape support directional visual transitions between landmarks.

\subsubsection{The Visual Corridors that Link Multiple Landmarks}

\begin{figure}[htbp]
    \centering
    \begin{subfigure}{\textwidth}
        \centering
        \includegraphics[width=\textwidth]{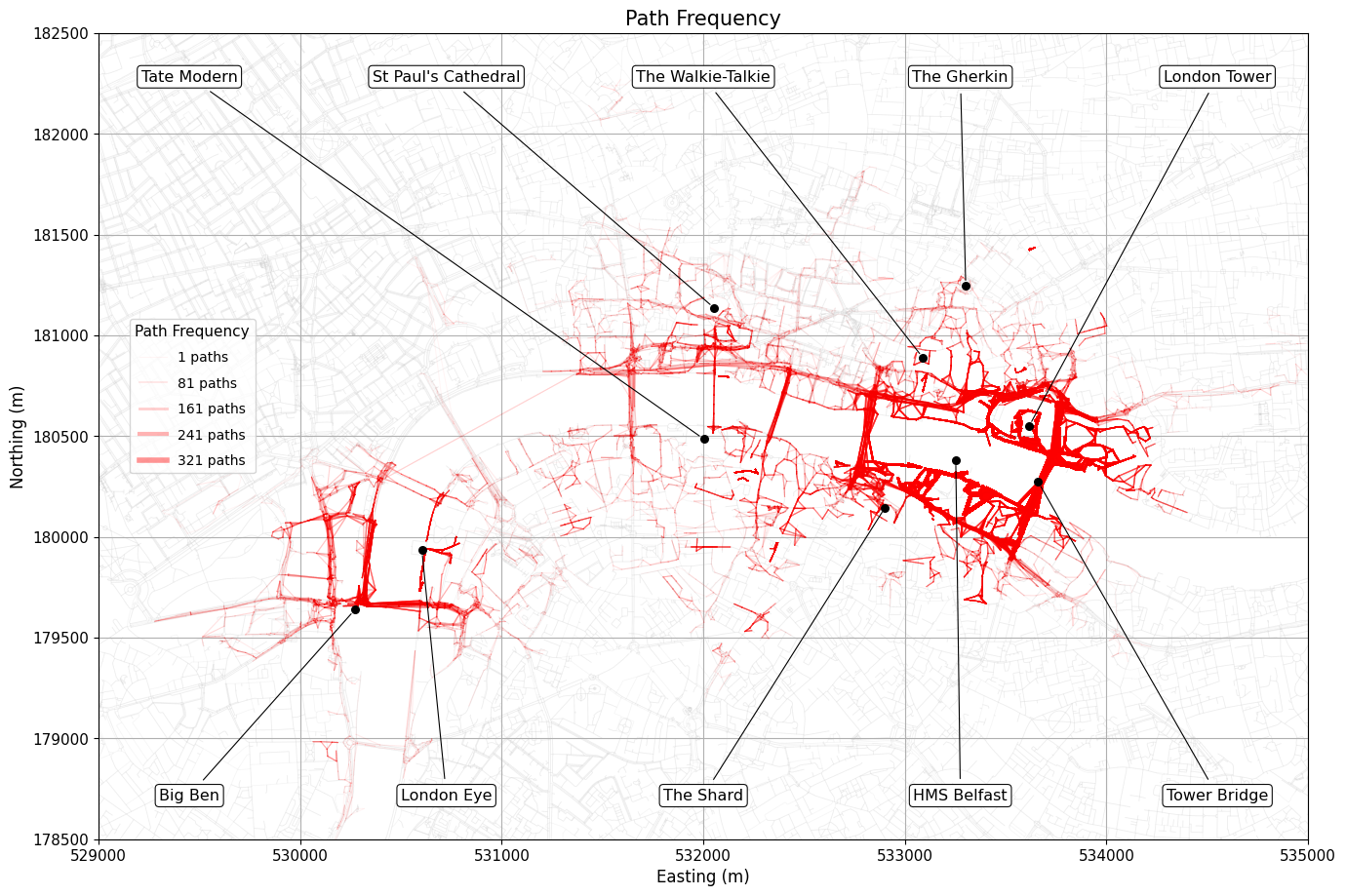}
        \caption{Accumulated VAV paths searched via a random walk experiment with 2000 walks, and a maximum step length of 80. Frequency of paths passing the same pair of SVIs is visualized.}
        \label{fig:total_paths}
    \end{subfigure}
    
    \begin{subfigure}{\textwidth}
        \centering
        \includegraphics[width=\textwidth]{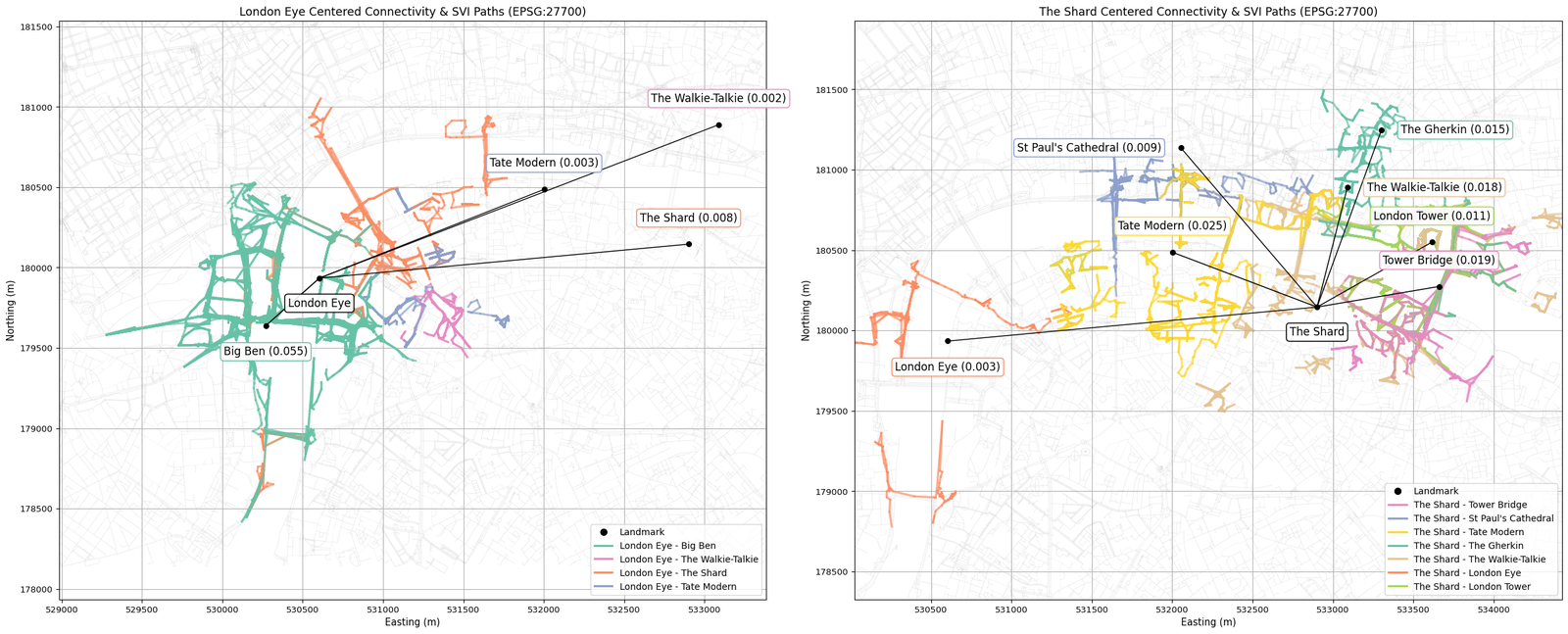}
        \caption{Accumulated VAV paths originating from the London Eye and The Shard, and the connectivity strength between landmark pairs. Only landmark pairs with linking strength over 0.002 are visualised.}
        \label{fig:specific_paths}
    \end{subfigure}

    \caption{Spatial distribution of VAV paths searched within a typical experiment based on random walk analysis. A VAV path is considered valid when its endpoints are SVI locations that reveal different landmarks.}
    \label{fig:paths_figures}
\end{figure}

As shown in Figure \ref{fig:total_paths}, the map illustrates the spatial distribution of total paths searched for ten landmarks. It was found that most of the paths concentrated around a circle on the east side of the case study area. The paths link landmarks such as The Walkie Talkie, London Tower, Tower Bridge, and The Shard, filling the surrounding space's riverside pedestrian walks and main roads.

Beyond the circular area, there are lower-level hot spots centred around St Paul's Cathedral and the Tate Modern, and around the London Eye and Big Ben. Strong and concentrated linear linking paths are observed between the St Paul's Cathedral–Tate Modern hot spot and the eastern circle on the northern bank. However, these paths are truncated at the western ward boundary of the City of London and do not extend further west. In contrast, linking on the southern bank is more diffuse. No single, strong, concentrated path is observed on the southern bank that links either Big Ben–London Eye or St Paul's Cathedral–Tate Modern with the eastern landmarks. Instead, multiple linking paths are evenly distributed on the road network in the wards of Bishop’s, Borough and Bankside, representing a more casual integration of landmarks into the visual background.

Figure \ref{fig:specific_paths} presents the linking paths originating from specific landmarks, the London Eye and The Shard, and the accumulated linking strength between origins and destinations. The London Eye is closely tied with Big Ben in visual perception, both through inter-visibility and visual co-existence. The longest VAV path can extend south along the River Thames, passing over Lambeth Bridge. It is also capable of establishing an intermediate visual connection between the London Eye and The Shard based on visual co-existence, although they are nearly on opposite sides of the southern bank.

Regarding The Shard, the landmark exhibits the strongest visual connection with the Tate Modern, which commonly serves as a foreground object when The Shard is observed in the background. This is followed by its strong link with The Gherkin and The Walkie Talkie, both high-rise landmarks. It is also noted that the visual interaction between The Shard and The Gherkin can occur at locations far from both landmarks, indicating their consistent roles as key elements of the urban skyline. Additionally, the linking strength between The Shard and St Paul's Cathedral, London Tower, and Tower Bridge is similar, nearly double that of the London Eye. For these traditionally recognised London landmarks along the River Thames, The Shard, as a relatively new construction, serves as a closely related background element in views of them.

\section{Discussion}

\subsection{An Image-based Framework for Visibility and Visual Relationship Analysis}
Our study shows that SVI can serve as a practical medium for sensing distant street-level space and capturing visual connections between landmarks. %
Table~\ref{tab:method_comparison} summarises its advantages and disadvantages compared to traditional 3D-based methods in visibility analysis. Compared with the open 3D visibility setup used in this study, the SVI-based method exhibits higher overall accuracy for detecting landmark visibility and maintains competitive performance as observation distance increases. %
The relative advantages of the SVI-based method come from two sources:  
Firstly, SVI generally presents better availability compared to high-quality 3D model data, especially in the urban realm. Though multiple openly available 3D model datasets exist, they are not always even in data quality or suitable for 3D visibility analysis in urban environments due to resolution or completeness limitations. Like the cases in Taipei and Dubai, height data presents relatively low completeness in the corresponding building dataset.
Secondly, SVI-based observation incorporates more real-world details, such as obstructions from trees, extended building roofs and billboards. %
Compared with a 3D method that simulates visibility primarily as a geometric intersection, the image-based method also relies on shape, texture, and other recognisable visual features of the target landmark, thereby reflecting image-based recognisability rather than geometry alone.

\begin{table}[htbp]
    \centering
     \scriptsize %
    \caption{A summary and comparison of advantages and disadvantages of different visibility analysis methods.}
    \label{tab:method_comparison}
    \begin{tabular}{%
        >{\raggedright\arraybackslash}p{1.5cm} %
        >{\raggedright\arraybackslash}p{5.5cm} %
        >{\raggedright\arraybackslash}p{5cm} %
    }
    \toprule
         & \textbf{3D-based Visibility} & \textbf{SVI-based Visibility} \\
    \midrule
    \textbf{Pros}
    & 1. Better spatial continuity of results\newline 2. Higher computation efficiency\newline 3. Flexibility in modifying models
    & 1. Higher analysis accuracy \newline 2. Aware of observation context\newline 3. Integration with urban road work \\
    \midrule
    \textbf{Cons}
    & 1. Model resolution and availability limits\newline 2. Incomplete consideration of semantic details and  other urban themes, e.g., vegetation
    & 1. Image resolution and metadata limits \newline 2. Limited availability beyond street space and in rural areas \\
     \midrule
    \textbf{Applications}
    & 1. Nature and urban environments\newline 2. Simulation and evaluation of the planning and construction impact
    & 1. Complex urban environments\newline 2. Evaluating the visual condition of existing construction\\
    \bottomrule
    \end{tabular}
\end{table}

Though with the advantages above, the application of the SVI-based method is restricted to the spatial and temporal availability of SVI data. The objects to observe are also fixed, which should be existing buildings or constructions. The SVI-based method can only be used to evaluate rather than simulate the visual impact. In these cases, the 3D-based method presents better flexibility. In addition, effective observation via SVI depends on the image resolution, metadata quality, and power of the CV model -- a similar condition also applies to observation based on well-functioning human eyes. Beyond a certain distance, or under poor lighting conditions, it is not feasible to recognise landmarks via both human and machine vision. In addition, seasonal bias is another common issue for both image-based and 3D-based methods in visibility analysis. Seasonal changes can substantially affect tree-related occlusion \citep{zhao_quantifying_2025} and therefore impact landmark visibility. For SVI-based methods, such changes may not be captured in a timely manner due to the relatively infrequent SVI updates. For 3D-based methods, seasonal effects are also difficult to represent because most urban 3D models provide limited information on tree morphology, canopy density, and vegetation type.

Overall, the image-based method provides a relatively low-cost complement to conventional LoS-based visibility analysis. It helps mitigate the tension between data availability, modelling complexity, and observational realism, particularly in complex urban environments where detailed 3D data are incomplete or unavailable.

\subsection{A New Gate to Comparative Urban Research and Heritage Conservation}

Our case studies, applying image-based visibility analysis to well-known landmark structures across global cities and to both modern and historic landmarks within London, demonstrate the method's potential for comparative urban research and heritage conservation.

From the first case study, different visibility patterns are recognised for iconic landmark structures in global cities.
For Petronas Towers in Kuala Lumpur and Burj Khalifa in Dubai, the landmarks are over 400 m in height and gain wide visibility from vast urban locations. 
The landmark visibility detected via image-based and 3D-based methods far exceeds the spatial extent of photo taking reflected by Flickr, resulting in a singular, radiating visual positioning. The landmarks serve not just as local attractions but as integral components of the broader visual background. 
In contrast, in cities such as London, Tokyo, and Paris, the visibility around selected landmarks is effectively confined within well-defined spatial boundaries, such as along riverbanks or principal urban axes and primary roads. Concerning the urban fabric, the landmarks exhibit a more harmonious visual positioning, contributing to the spatial structure and distinctive characteristics of local areas. %
With the visibility patterns revealed above, our research provides a novel alternative for evaluating the spatial impact of globally recognised landmarks. It offers further opportunities to improve the design quality and vitality of landmark-related space from a planning perspective.

In the second case study, our method effectively captures the complex landscape patterns by multiple urban landmarks, a critical aspect for heritage preservation. 
By combining visibility and accessibility via VAV paths and employing the random-walk algorithm, we examine the mutual relationships among London landmarks. We can discern how urban landmarks that do not interact visually directly, such as Tower Bridge and the London Eye, establish indirect connections through unexpected urban spaces. This analysis not only quantifies overall connection strength but also identifies the key locations and corridors that facilitate these links, which have been shown to play positive roles in human way-finding tasks \citep{omer_implications_2007}. 
As an extended exploration, we found that bridges on the River Thames serve as vital spatial and visual corridors for linking the landmarks. As shown in Table \ref{tab:bridge_traffic}, over 31\% of the total paths pass through the nine bridges over the River Thames. Specifically, if London Bridge were removed from the River Thames, 5.46\% of the current VAV paths would be cut off, significantly reducing the linking between Tower Bridge, The Walkie Talkie, The Shard, and Tate Modern. 

Urban authorities often seek to preserve long-established visual relationships with landmarks, such as visibility from designated viewpoints and directions \citep{tavernor_visual_2007}. For example, the London View Management Framework has established eight protected vistas of St. Paul's Cathedral, imposing height restrictions on buildings that might obstruct these landmarks \citep{greater_london_authority_london_2012}. 
Compared with traditional heritage preservation practices, the proposed image-based visibility method helps identify everyday street visual corridors that are not necessarily covered by existing protection frameworks, but mediate important visual connections among landmarks. This approach thus shows promise for Heritage Impact Assessment (HIA), especially in dynamic urban environments where visual relationships may be reshaped by ongoing development \citep{ashrafi_heritage_2021}. More broadly, this work extends the exploration by \citet{shen_functional_2022} and \citet{natapov_can_2013} on semantics-enriched visibility graphs.

\subsection{Limitations and Future Work}
There are several limitations in this study.
To achieve a trade-off between performance and memory usage, our study prevents the usage of over-complicated Visual-Language Models (VLMs), such as GPT 4o \citep{openai_hello_2024}, Qwen-VL \citep{bai_qwen-vl_2023}, CogVLM \citep{wang_cogvlm_2024}. Integrated with multi-modal capabilities from Large Language Models (LLMs), these models generally achieve better performance in zero-shot detection and also advance on other tasks such as Visual Question Answering (VQA) and image captioning. However, the additional functionality dramatically increases the model size and the cost for calling and inference, which is unnecessary for elaborating the proposed image-based visibility method. Nevertheless, more advanced VLMs may improve the ability to identify a broader range of target objects beyond iconic landmarks, although such generalisation remains to be empirically validated. Similarly, due to the scope limitation of this study, we did not systematically test the proposed method under challenging visual conditions, such as fog, rain, or temporary obstructions. These factors may reduce detection recall. Future work could evaluate how these conditions affect the robustness of street-view-based long-distance urban observation.

Additionally, our study only compared the performance of the image-based visibility analysis method with visibility simulated from open 3D urban models at overlapping locations available to both analytical methods. We acknowledge that this comparison may not be perfect. First, it is difficult to retrieve high-quality urban 3D models with both comprehensive spatial coverage and the necessary representation of terrain elevation, tree canopies, and accurate building heights. Most complete datasets of this kind are from closed-source or commercial data providers. Second, our comparison may place the 3D-based method in a disadvantaged position, as there may be other locations detected as visible by the 3D method but not included in the validation dataset due to the absence of SVI data. Even so, the initial aim of this study is to investigate the potential of image-based method for visibility analysis, rather than to prove its superiority over 3D-based methods.

Moreover, limited to the scope of the study, the usability exploration of the visibility graph, specifically integrating visibility into a road-based spatial network, is not sufficient. 
Our study reveals that visibility often represents a hyper-space relationship, which naturally complements the classical spatial model defined on spatial proximity and fits well with the graph structure. 
As a future direction, we can apply the relationship in use cases such as building height prediction, geo-localisation, urban environment embedding, and utilising graph deep learning methods.

\section{Conclusion}
By combining SVI with zero-shot object detection, this study proposes a visibility analysis method for recognisable urban landmarks and creates opportunities to analyse broader visual and spatial contexts of observation.
Further, a heterogeneous visibility graph is constructed to address the interactions among visual objects and quantify their connection strength.
Using two case studies to investigate the visibility and visual relationships of urban landmarks, the study demonstrates the method's reliability in the present cases and its potential application in urban design and heritage conservation. 
Specifically, the image-based visibility analysis helps address limitations commonly encountered in open 3D-based visibility workflows, such as restricted availability of high-resolution 3D data and ignoring the street-level visual context. 
Furthermore, for urban planners, the proposed visibility graph showcases the potential of revealing where landmarks are recognisable from everyday street environments, supports the assessment of visual corridors and skylines,  and can complement existing heritage conservation practices that focus on fixed protected viewpoints. %

\section*{Acknowledgments}
We express our gratitude to the members of the NUS Urban Analytics Lab for their valuable discussions and insights, and to Shuhui Wu for her assistance with the illustration.
The first author is supported by the National University of Singapore under the President’s Graduate Fellowship.
This research has been supported by Takenaka Corporation.
This research is part of the project Large-scale 3D Geospatial Data for Urban Analytics, which is supported by the National University of Singapore under the Start Up Grant R-295-000-171-133.
This research is part of the project Multi-scale Digital Twins for the Urban Environment: From Heartbeats to Cities, which is supported by the Singapore Ministry of Education Academic Research Fund Tier 1.

\section*{Author contributions}

\textbf{Zicheng Fan:}
Conceptualisation;
Methodology; 
Software;
Validation;
Formal analysis;
Investigation;
Data Curation;
Writing - Original Draft;
Visualisation.

\textbf{Kunihiko Fujiwara:}
Formal analysis;
Investigation;
Data Curation;
Writing - Original Draft;
Writing - Review \& Editing;
Visualisation.

\textbf{Pengyuan Liu:}
Methodology; 
Writing - Review \& Editing;
Supervision;

\textbf{Fan Zhang:}
Writing - Review \& Editing;

\textbf{Filip Biljecki:}
Conceptualisation;
Investigation;
Methodology;
Writing - Review \& Editing;
Visualisation;
Supervision;
Project administration;
Funding acquisition.

\section*{Declaration of generative AI and AI-assisted technologies in the writing process}

During the preparation of this work the authors used ChatGPT in order to proofread the text. After using this tool, the authors reviewed and edited the content as needed and take full responsibility for the content of the publication.

\newpage

\appendix
\section{Integrating SVI Locations into Graph with Road Connection}
\label{sec:sample:appendix_A}

This section provides additional details on how SVI locations are integrated into a road-based graph. Because SVI points are not always evenly distributed along road segments, virtual nodes are introduced where necessary to preserve network connectivity. 

\begin{figure}[htbp]
    \centering
    \begin{subfigure}{1\textwidth}
        \centering
        \includegraphics[width=\textwidth]{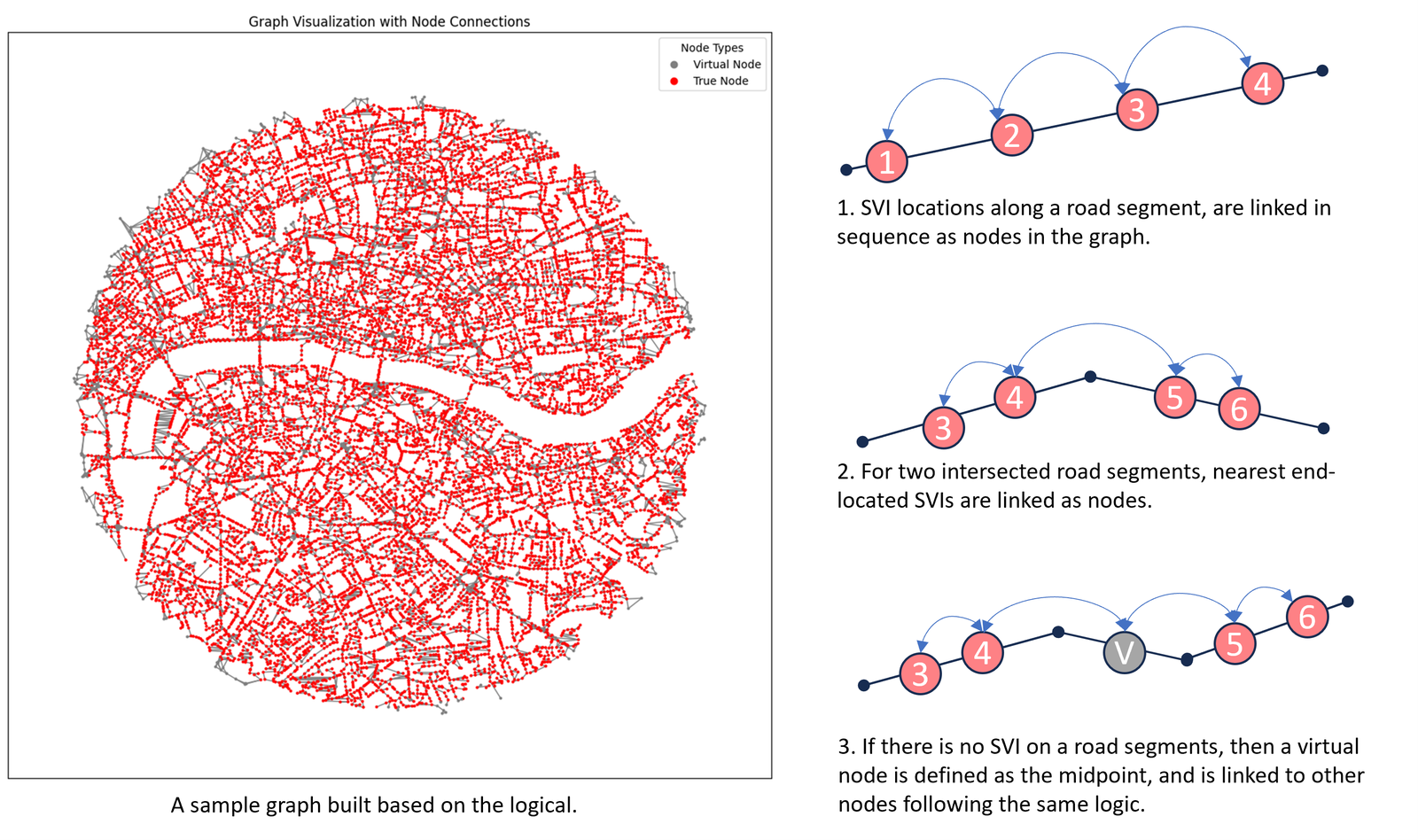}
    \end{subfigure}
    \caption{Right: Steps for connecting SVI locations into a graph based on spatial proximity and road network connections. Virtual nodes are defined to maintain network connection when SVI data is not available on a road segment. Left: A sample graph built based on the logic in London, centring on The Shard.}
    \label{fig:svi_nodes}
\end{figure}

\newpage
\section{Positioning of Landmarks on SVI}
\label{sec:landmark_position}

This section details the geometric procedure used to locate and crop the target landmark within panoramic SVI. 

\paragraph{Calculate Relative Position and Angle}

Let a panoramic SVI be at location $O=(x_{o},y_{o})$, the landmark at location $L=(x_{\ell},y_{\ell})$, both in the local projected Coordinate Reference System (CRS). The panorama has width $W$ and height $H_{\mathrm{img}}$ in pixels, and heading $h$, clockwise from geographic north in radians. The landmark has a height $H$ in meters. %
The Euclidean distance from the observer to the landmark is:

\begin{equation}
d = \|L-O\|_2
  \;=\; \sqrt{(x_{\ell}-x_{o})^{2} + (y_{\ell}-y_{o})^{2}}.
\label{eq:distance}
\end{equation}

Using the planar axes convention (east $x$, north $y$), we calculate the horizontal azimuth of the landmark relative to the observer from north as:

\begin{equation}
\alpha = \operatorname{atan2}\bigl(x_{\ell}-x_{o},\,y_{\ell}-y_{o}\bigr)
\quad[\text{radians}].
\label{eq:azimuth}
\end{equation}

By converting both the panorama heading and landmark horizontal azimuth to degrees, we can calculate their difference and wrap to $0^{\circ}$–$360^{\circ}$. This quantity tells us how far to the right ($0^{\circ}$–$180^{\circ}$) or
left ($180^{\circ}$–$360^{\circ}$) of the heading direction, the landmark lies in panoramic SVI.

\begin{equation}
\Delta\alpha^{\circ}
  = \bigl(\alpha^{\circ} - h^{\circ}\bigr) \bmod 360.
\label{eq:delta_angle}
\end{equation}

Because the panorama is commonly stored in an equirectangular projection, the horizontal angle is linear in pixel column. With the horizontal angle difference $\Delta\alpha^{\circ}$, we can further calculate the horizontal pixel coordinate of the landmark on a panorama.

\begin{equation}
x_{\text{pix}}
  = \biggl(\frac{W}{2} + \frac{\Delta\alpha^{\circ}}{360^{\circ}}\,W\biggr)
    \bmod W.
\label{eq:xp}
\end{equation}

\newpage
\paragraph{Zoom-in SVI Towards Landmarks}
The zoom-in extent is determined based on the distance $d$ between the SVI location and the landmark and the landmark's physical height $H$. Specifically, the landmark's elevation angle in an image is: 

\begin{equation}
\epsilon = \arctan\Bigl(\tfrac{H}{d}\Bigr).
\label{eq:elev}
\end{equation}

Assuming a full vertical field of view
$\theta_v = 180^{\circ}$ and panorama height $H_{\mathrm{img}}$ (px), hence the landmark’s height in pixels is:

\begin{equation}
h_{\text{pix}}
  \;=\;
  \epsilon\;
  \frac{H_{\mathrm{img}}}{\theta_v}.
\label{eq:h_pix}
\end{equation}

Taking the horizon line (image mid-height) as the reference, the bottom and top of the extent are defined to focus on the upper half of the landmark. This framing is a heuristic designed for tall landmarks whose upper portions are more likely to remain recognisable under partial occlusion:

\begin{equation}
y_{\text{bottom}} \;=\; \frac{H_{\mathrm{img}}}{2} \;-\; 0.50\,h_{\text{pix}},
\qquad
y_{\text{top}}    \;=\; \frac{H_{\mathrm{img}}}{2} \;-\; 1.00\,h_{\text{pix}}.
\label{eq:y_bounds2}
\end{equation}

Let $x_{\text{pix}}$ be the landmark’s central column obtained from the azimuth difference.  
Then the horizontal limits can be defined symmetrically:

\begin{equation}
x_{\text{left}}  \;=\; x_{\text{pix}} - \frac{y_{\text{bottom}}-y_{\text{top}}}{2},
\qquad
x_{\text{right}} \;=\; x_{\text{pix}} + \frac{y_{\text{bottom}}-y_{\text{top}}}{2}.
\label{eq:x_bounds2}
\end{equation}

The four bounds define a scale-consistent, nearly square zoom that adapts to landmark height and distance.
\begin{equation}
B \;=\;
\Bigl[
  x_{\text{left}},\;
  y_{\text{top}},\;
  x_{\text{right}},\;
  y_{\text{bottom}}
\Bigr].
\label{eq:crop2}
\end{equation}

\newpage

\section{SVI Metadata and 3D Data Sources}
\label{sec:sample:appendix_C}

This section reports the data sources used to construct the 3D visibility simulation and the spatial and temporal distribution of the collected SVI. 

Regarding the spatial distribution of SVI, Dubai presents relatively limited coverage along the road network in the western, southern, and northeastern parts of the study area, mainly due to insufficient image collection by Google Street View. For the other cities, the spatial coverage of SVI data is generally sufficient. Dubai, Kuala Lumpur, and Taipei are located in desert, tropical, and subtropical climate regions, respectively, where evergreen vegetation and relatively stable greenery conditions are expected. For the remaining cities, we consider April to October as the period with relatively stable street greenery. In Paris, Tokyo, and London, 84.10\%, 87.76\%, and 93.47\% of the collected images fall within this period, respectively. We acknowledge that, in Paris, images collected during the winter season still account for a non-negligible proportion, which may introduce variation into landmark visibility detection. However, a systematic evaluation of this impact is beyond the scope of the present study. Future work will further quantify the potential influence of seasonal variation on image-based visibility detection.

\begin{table}[htbp]
\centering
\caption{Data sources for 3D city model generation with VoxCity. "Base" and "Comp." represent base and complementary data for building footprint and height. Complementary data is used to complement missing building height values in the base data.}
\scriptsize
\begin{tabular}{p{0.15\textwidth}p{0.25\textwidth}p{0.25\textwidth}p{0.25\textwidth}}
\hline
Landmark & Building Footprint and Height & Tree Canopy Height & Terrain Elevation \\
\hline
The Shard & Base: OSM, Comp.: England 1m Composite DTM/DSM \citep{envagency2024dtm, envagency2024dsm} & High Resolution 1m Global Canopy Height Maps \citep{Tolan2024-an} & England 1m Composite DTM \citep{envagency2024dtm} \\
\hline
Petronas Towers & Base: OSM, Comp.: Open Building 2.5D Temporal \citep{Sirko2023-hb} & High Resolution 1m Global Canopy Height Maps \citep{Tolan2024-an} & FABDEM \citep{Hawker2022-by} \\
\hline
Tokyo Tower & Base: OSM, Comp.: UT-GLOBUS \citep{Kamath2024-at} & High Resolution 1m Global Canopy Height Maps \citep{Tolan2024-an} & FABDEM \\
\hline
Eiffel Tower & EUBUCCO \citep{Milojevic-Dupont2023-bw} & High Resolution 1m Global Canopy Height Maps \citep{Tolan2024-an} & RGE ALTI \citep{rgealti2024} \\
\hline
Taipei 101 & Base: OSM, Comp.: None & High Resolution 1m Global Canopy Height Maps \citep{Tolan2024-an} & FABDEM \\
\hline
Burj Khalifa & Base: OSM, Comp.: None & High Resolution 1m Global Canopy Height Maps \citep{Tolan2024-an} & FABDEM \\
\hline
\end{tabular}%

\label{tab:voxcity_source_vertical}
\end{table}

\newpage
\begin{figure}[htbp]
    \centering
    \begin{subfigure}{1\textwidth}
        \centering
        \includegraphics[width=1\textwidth]{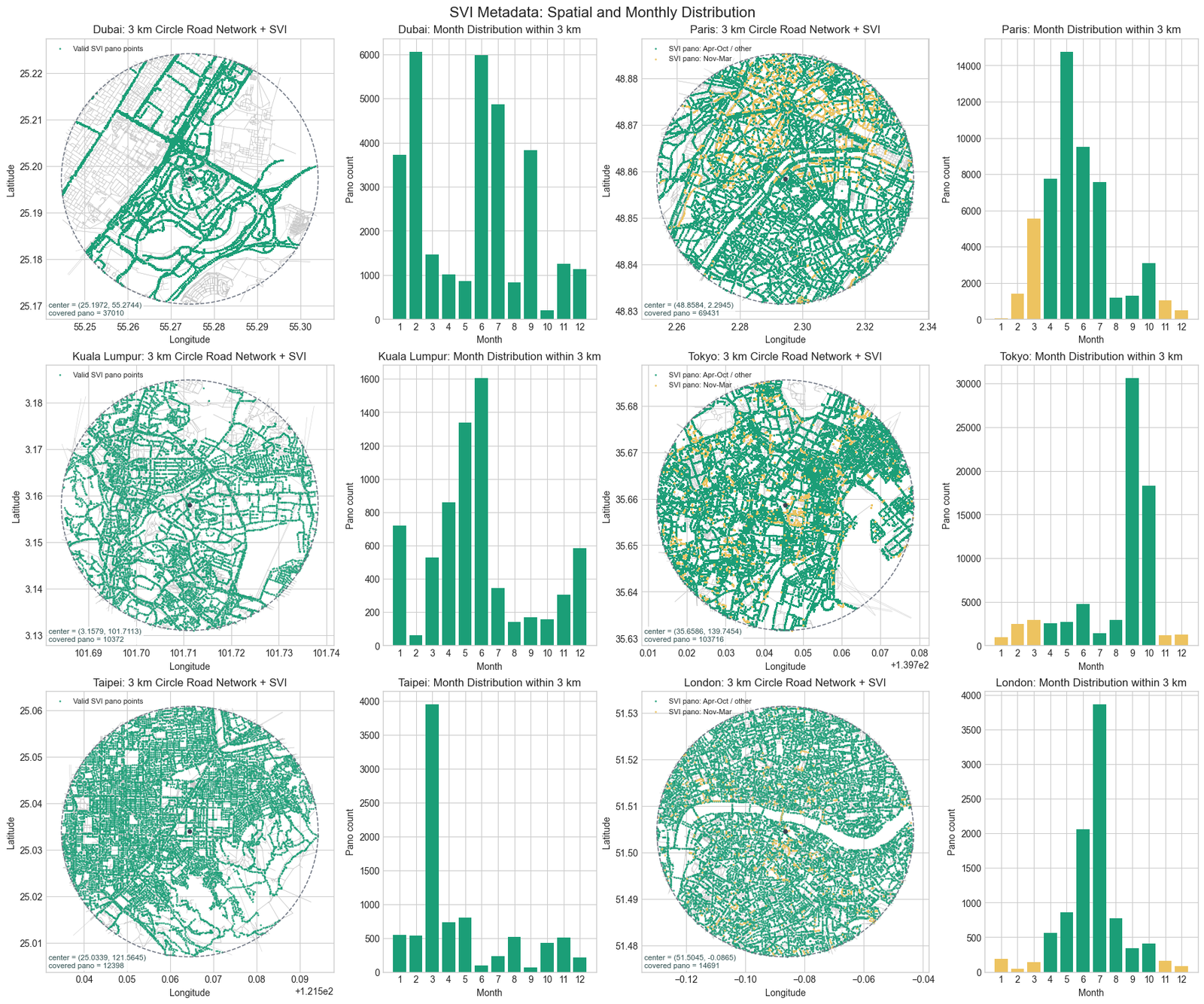}
    \end{subfigure}
    \caption{Spatial and monthly distribution of collected SVIs across six global cities.}
    \label{svi_metadata}
\end{figure}

\newpage
\section{Landmark Visibility Validation Dataset}
\label{sec:sample:appendix_B}

\begin{table}[ht]
    \centering
    \caption{Components of validation set for case study 1. For each landmark, the validation set is created by randomly sampling 100 SVIs from distance bands of 0-500 m, 500-1000 m, 1000-1500 m and 1500 m+. The images are then manually labelled as 'visible' and 'invisible' for the corresponding landmark, using Label Studio \citep{tkachenko_label_2020}. }
    \scriptsize
        \begin{tabular}{p{0.15\textwidth}p{0.2\textwidth}p{0.2\textwidth}p{0.2\textwidth}}
        \toprule
        \textbf{Tower} & \textbf{Landmark Visible} & \textbf{Landmark Invisible} & \textbf{Total Images}\\
        \midrule
        Eiffel Tower & 86 & 314 & 400\\
        Tokyo Tower & 39 & 361 & 400\\
        Petronas Towers & 97 & 303 & 400\\
        The Shard & 73 & 327 & 400\\
        Taipei 101 & 103 & 297 & 400\\
        Burj Khalifa & 195 & 205 & 400\\
        \bottomrule
    \end{tabular}
    \label{tab:towers_visibility}
\end{table}

\begin{figure}[htbp]
    \centering
    \begin{subfigure}{1\textwidth}
        \centering
        \includegraphics[width=0.85\textwidth]{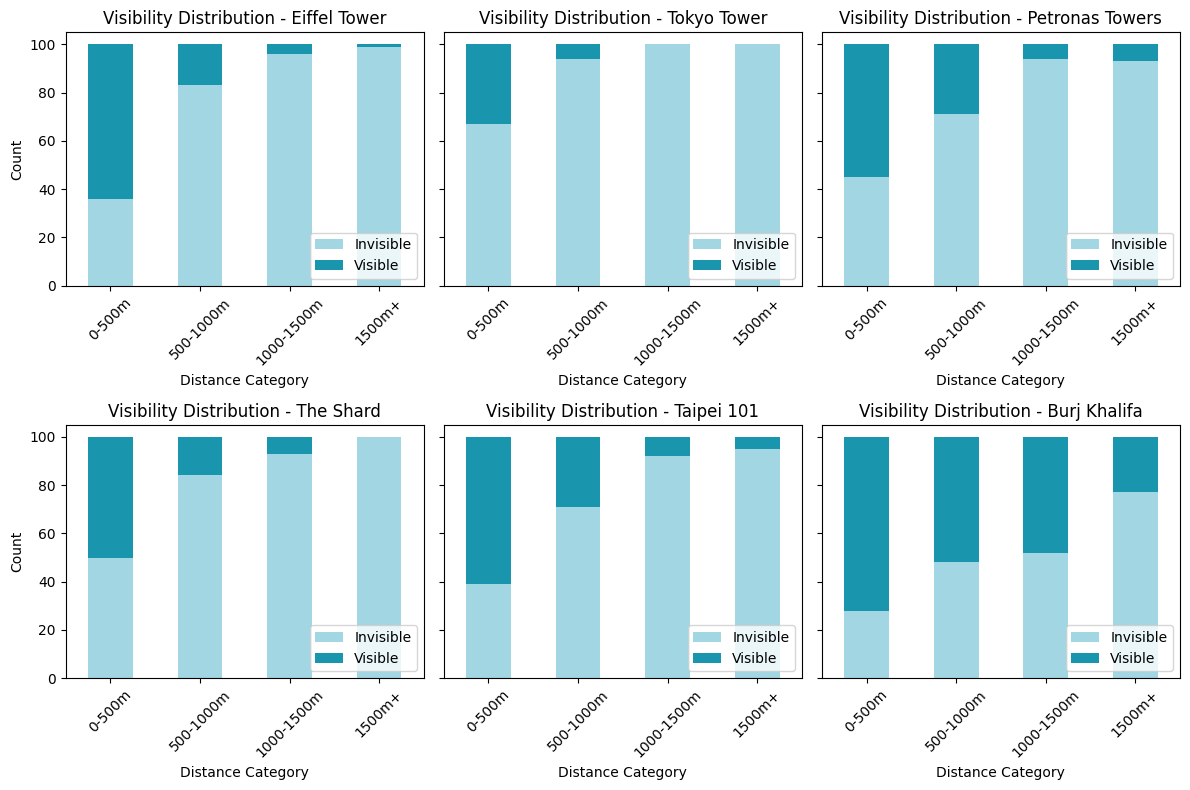}
    \end{subfigure}
    \caption{Distribution of visible and invisible samples within each distance band in the validation set.}
    \label{validation_distribution}
\end{figure}

\newpage
\section{Ablation Experiments}
This section provides ablation results to investigate: (1) the impact of zoom-in crop strategy on the image-based landmark visibility detection. (2) difference between imagery and text as query for visibility detection. It is turned out that the zoom-in preprocessing can significantly improve the performance of zero-shot object detection. Addtionally, image queries present comparative advantages over text queries for detecting both visible and invisible classes. The advantage is prominent for Tokyo Tower, The Shard, Petronas Towers, and Buij Khalifa. For the remaining two landmarks, however, text queries achieve relatively better performance than image queries.

\begin{table}[h]
    \scriptsize
    \centering
    \caption{Ablation results for different image inputs and query settings in image-based landmark visibility detection.}
    \begin{tabular}{lccccccc}
        \toprule
        & & \multicolumn{3}{c}{Visible Class} & \multicolumn{3}{c}{Invisible Class} \\
        \cmidrule(lr){3-5} \cmidrule(lr){6-8}
        Case & Accuracy & Precision & Recall & F1-score & Precision & Recall & F1-score \\
        \midrule
        Full Panorama + Image Query & 0.4542 & 0.2735 & 0.7302 & 0.3980 & 0.8042 & 0.3636 & 0.5008 \\
        Upper Half + Image Query & 0.4100 & 0.2687 & 0.8061 & 0.4030 & 0.8148 & 0.2800 & 0.4168 \\
        Zoom-in Crop + Image Query & 0.8712 & 0.6797 & 0.9056 & 0.7766 & 0.9652 & 0.8600 & 0.9096 \\
        Zoom-in Crop + Text Query & 0.8338 & 0.6283 & 0.8010 & 0.7042 & 0.9282 & 0.8445 & 0.8844 \\
        \bottomrule
    \end{tabular}
    \label{tab:visibility_ablation}
\end{table}

\begin{figure}[htbp]
    \centering
    \begin{subfigure}{1\textwidth}
        \centering
        \includegraphics[width=1\textwidth]{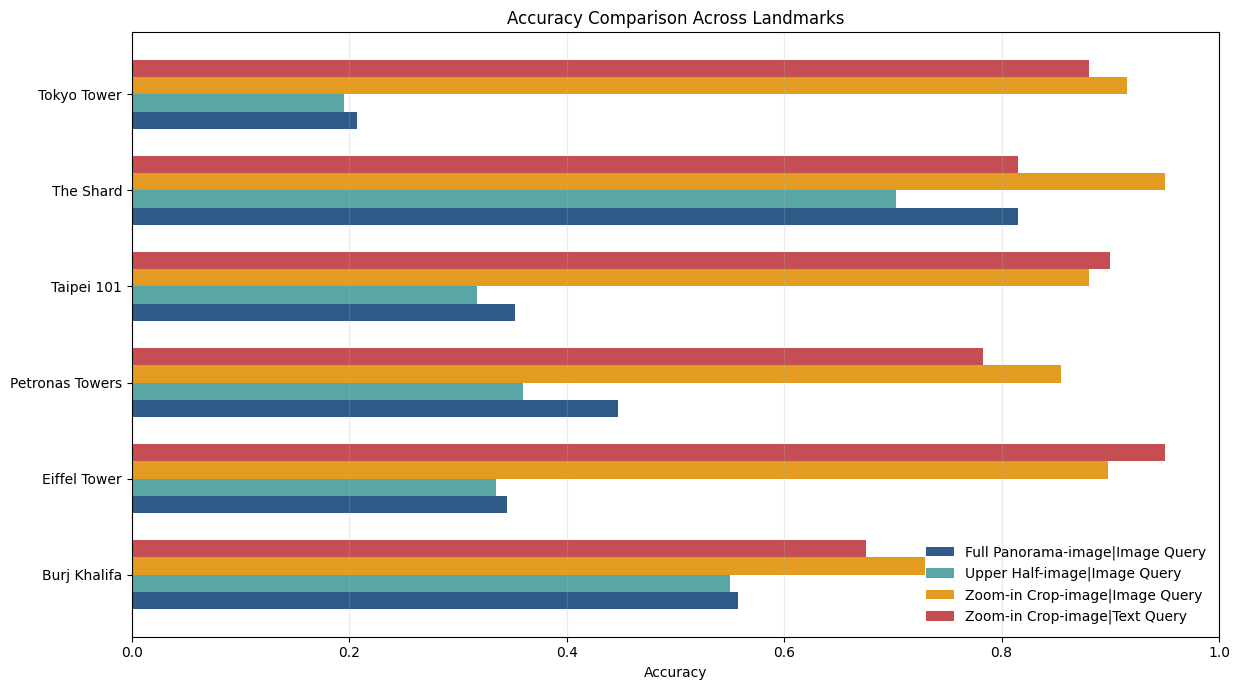}
    \end{subfigure}
    \caption{Ablation results across six global landmarks.}
    \label{fig:ablation_six_landmarks}
\end{figure}

\newpage
\section{Results Comparison between SVI and 3D Methods}
\label{sec:sample:appendix_D}

This section provides supplementary results comparing the SVI-based and 3D-based visibility analyses. The raw spatial distributions of visible locations and spatial grids are presented to complement the aggregated results in the main text. Additional balanced-accuracy results are also included to account for the influence of class imbalance in the validation dataset.
\vspace{-1em}
\begin{figure}[htbp]
    \centering
    \begin{subfigure}{\textwidth}
        \centering
        \includegraphics[width=0.85\textwidth]{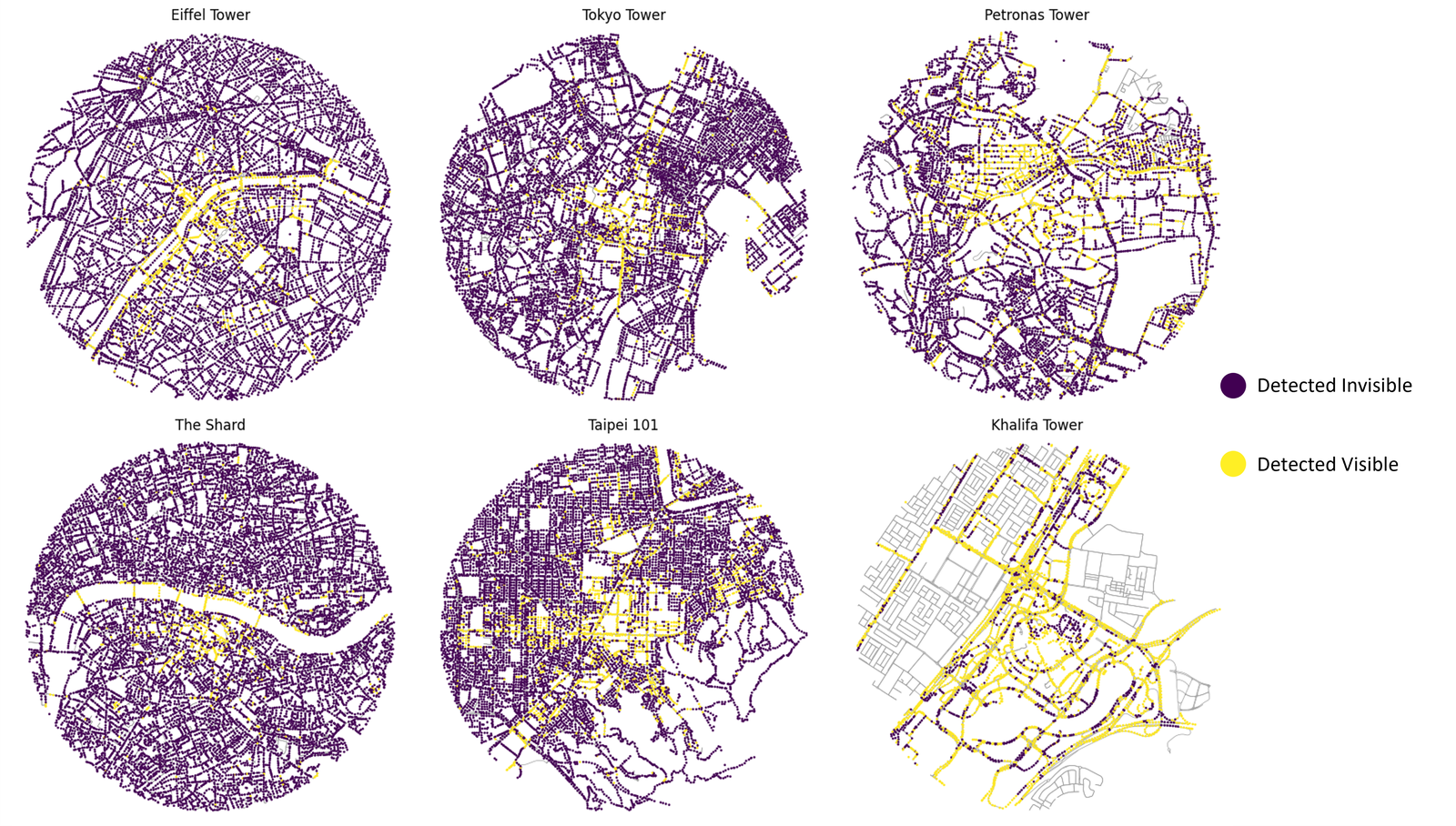}
        \caption{The spatial distribution of image-based visibility detection results.}
    \end{subfigure}
    \vspace{-0.5em}
    \begin{subfigure}{\textwidth}
        \centering
        \includegraphics[width=0.85\textwidth]{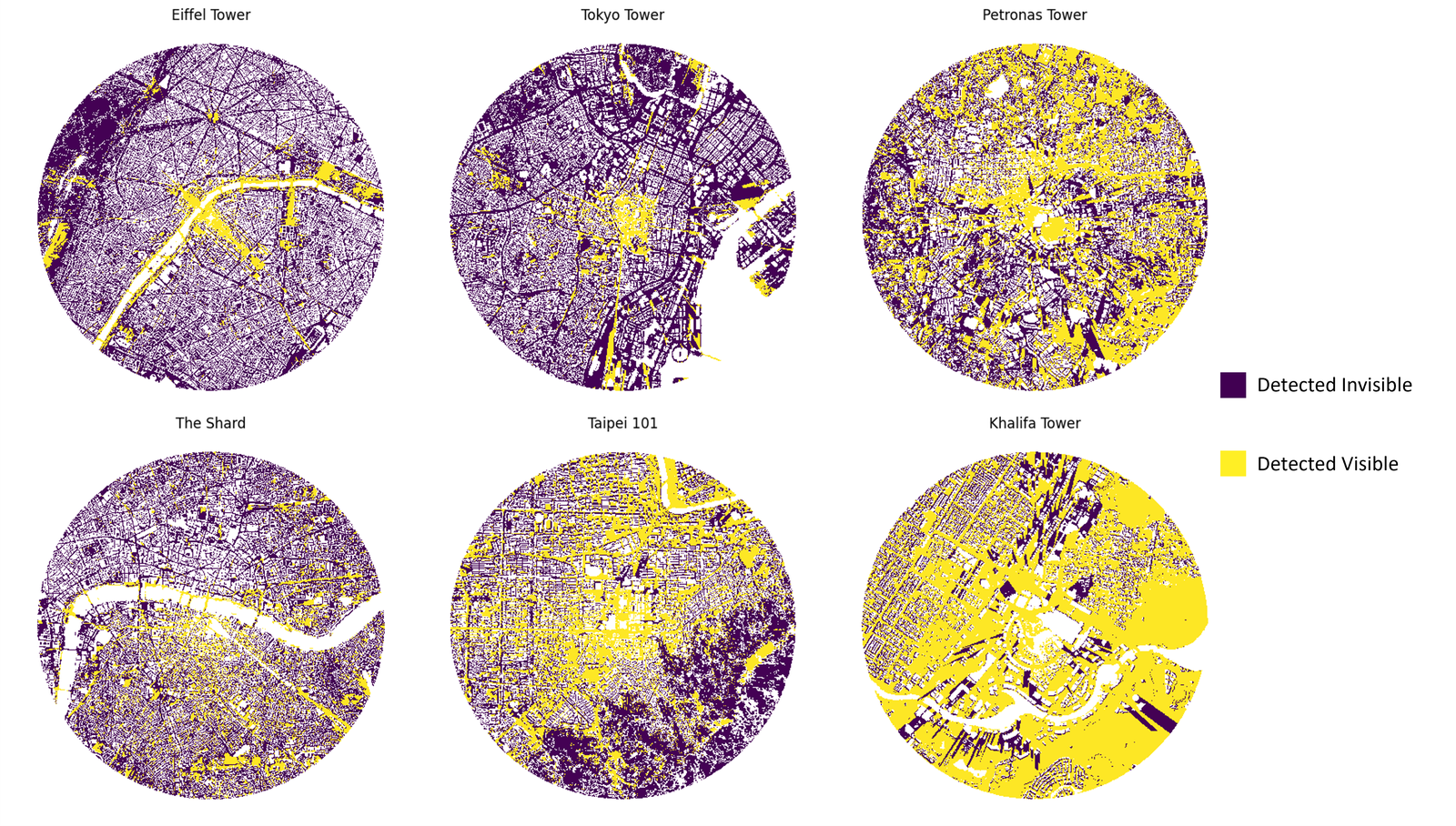}
        \caption{The spatial distribution of 3D-simulated visibility results based on open data sources.}
    \end{subfigure}
    \caption{Maps illustrating the raw results from SVI-based and 3D-based visibility analyses.}
    \label{fig:raw_visibility_map}
    
\end{figure}

\begin{figure}[htbp]
    \centering

    \begin{subfigure}{\textwidth}
        \centering
        \includegraphics[width=\textwidth]{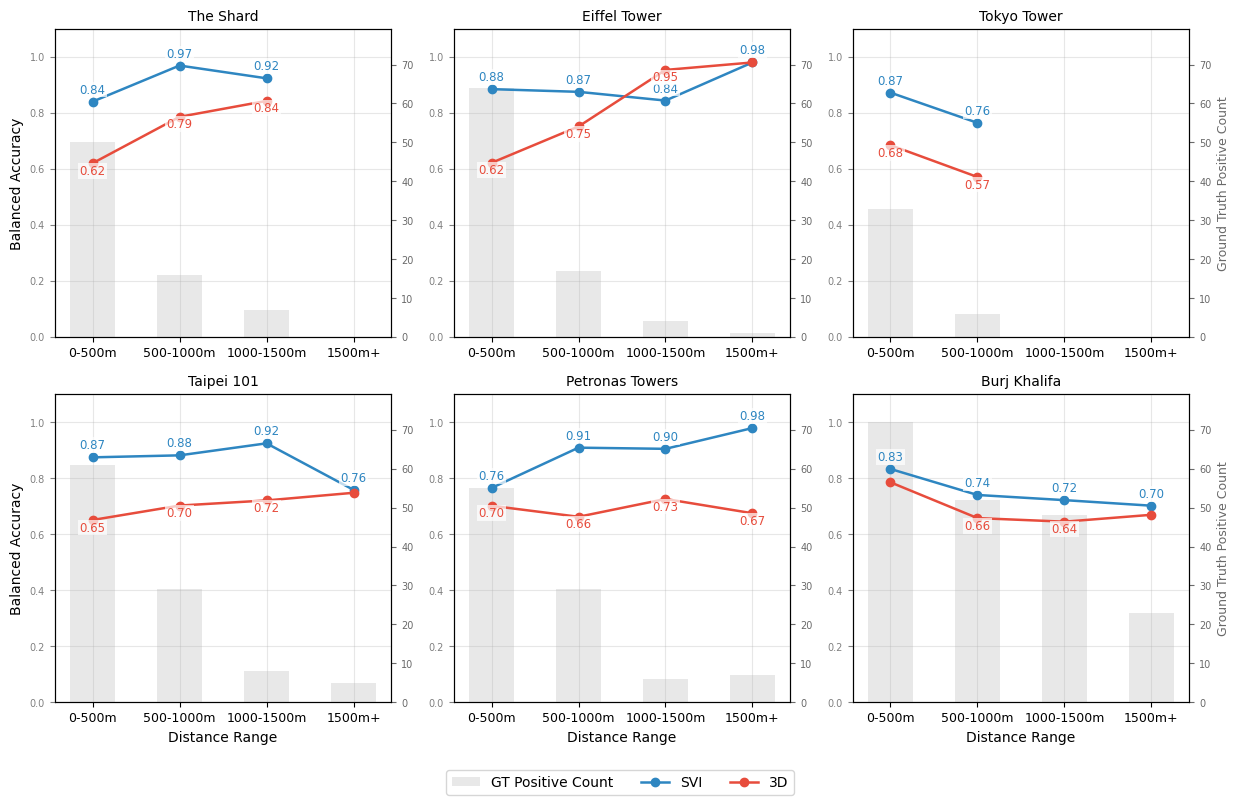}
        \label{fig:recall_lines}
    \end{subfigure}

    \caption{Line plots comparing the balanced accuracy of landmark visibility detection based on image and 3D methods.}
    \label{fig:additional_line_plots}
\end{figure}

\newpage
\section{Implementation Details for Random Walk Simulation}
\label{sec:random_walk}
This section describes the implementation details of the random-walk-based VAV path simulation. 
\paragraph{Angle-penalised random walk}
At each step of the random walk, let the current street-view location be \(s_t\), the previous street-view location be \(s_{t-1}\), and the set of candidate neighbouring locations be \(\mathcal{N}(s_t)\). When a previous movement direction is available, we compute the incoming direction vector
\begin{equation}
\mathbf{v}_1 = \mathbf{x}(s_t) - \mathbf{x}(s_{t-1}),
\end{equation}
and, for each candidate neighbour \(s' \in \mathcal{N}(s_t)\), the outgoing candidate direction vector
\begin{equation}
\mathbf{v}_2(s') = \mathbf{x}(s') - \mathbf{x}(s_t),
\end{equation}
where \(\mathbf{x}(\cdot)\) denotes the planar coordinates of an SVI location.

The directional consistency between the previous and candidate movement is measured using cosine similarity,
\begin{equation}
\cos \theta(s') =
\frac{\mathbf{v}_1 \cdot \mathbf{v}_2(s')}
{\|\mathbf{v}_1\| \, \|\mathbf{v}_2(s')\|}.
\end{equation}
In the implementation, if either \(\|\mathbf{v}_1\|=0\) or \(\|\mathbf{v}_2(s')\|=0\), the cosine value is set to \(1.0\).

A simple angle penalty is then applied:
\begin{equation}
w(s') =
\begin{cases}
0.1, & \text{if } \cos \theta(s') < 0,\\
1.0, & \text{if } \cos \theta(s') \geq 0.
\end{cases}
\end{equation}
Thus, candidate moves implying a backward turn (i.e., an angle greater than \(90^\circ\)) are down-weighted by a factor of \(0.1\), while forward or lateral moves retain full weight. The next location is sampled according to the normalized probability
\begin{equation}
P(s' \mid s_t, s_{t-1}) =
\frac{w(s')}
{\sum\limits_{u \in \mathcal{N}(s_t)} w(u)}.
\end{equation}
This strategy discourages U-turns and unrealistic reversals while preserving stochastic exploration of the road network.

\paragraph{Matrix correlation}
For each parameter setting and random seed, the random walk simulation produces a landmark-to-landmark connectivity matrix
\(\mathbf{M} \in \mathbb{R}^{n \times n}\),
where \(M_{ij}\) denotes the normalized directional connectivity strength from landmark \(i\) to landmark \(j\).
To compare two matrices \(\mathbf{M}^{(a)}\) and \(\mathbf{M}^{(b)}\), the diagonal entries are excluded and the remaining off-diagonal elements are flattened into vectors
\begin{equation}
\mathbf{m}^{(a)} = \mathrm{vec}_{\mathrm{off}}(\mathbf{M}^{(a)}),
\qquad
\mathbf{m}^{(b)} = \mathrm{vec}_{\mathrm{off}}(\mathbf{M}^{(b)}).
\end{equation}
Their similarity is then measured using the Pearson correlation coefficient. This metric evaluates whether two runs preserve a similar overall pattern of directional landmark relations, independent of absolute scale.

\paragraph{Relative Frobenius norm}
To quantify absolute matrix-level differences, we compute the relative Frobenius norm between two connectivity matrices,
\begin{equation}
d_F\!\left(\mathbf{M}^{(a)}, \mathbf{M}^{(b)}\right)
=
\frac{\left\| \mathbf{M}^{(a)} - \mathbf{M}^{(b)} \right\|_F}
{\left\| \mathbf{M}^{(a)} \right\|_F},
\end{equation}
where
\begin{equation}
\|\mathbf{A}\|_F
=
\sqrt{\sum_{i=1}^{n}\sum_{j=1}^{n} A_{ij}^2}.
\end{equation}
This metric measures the magnitude of discrepancy between two runs relative to the baseline matrix, and is sensitive to changes in connection strength values.

\paragraph{Top-\(k\) edge Jaccard similarity}
To evaluate the reproducibility of the strongest directional relations, we extract the set of top-\(k\) directed edges from each matrix after excluding self-connections. Let
\begin{equation}
\mathcal{E}_k(\mathbf{M})
\end{equation}
denote the set of the \(k\) highest-weight ordered landmark pairs \((i,j)\), \(i \neq j\), in matrix \(\mathbf{M}\).
For two runs, the overlap of their strongest edges is measured using Jaccard similarity,
\begin{equation}
J_k\!\left(\mathbf{M}^{(a)}, \mathbf{M}^{(b)}\right)
=
\frac{\left| \mathcal{E}_k(\mathbf{M}^{(a)}) \cap \mathcal{E}_k(\mathbf{M}^{(b)}) \right|}
{\left| \mathcal{E}_k(\mathbf{M}^{(a)}) \cup \mathcal{E}_k(\mathbf{M}^{(b)}) \right|}.
\end{equation}
A higher value indicates that the most salient directional landmark connections are more consistently reproduced across repeated simulations.

\newpage

\section{Bridge-Passing Visibility-Accessibility-Visibility (VAV) Paths}
\label{sec:sample:appendix_F}

\begin{table}[!htbp]
    \centering
    \captionsetup{skip=4pt}

    \caption{VAV paths identified that pass over bridges across the River Thames within the case study area. A total of 9 bridges are associated with the identified VAV paths. Among them, Tower Bridge is the most prominent, supporting the largest number of VAV paths and serving as both a landmark and a viewpoint.}
    \label{tab:bridge_traffic}

    \scriptsize
    \setlength{\tabcolsep}{5pt}
    \renewcommand{\arraystretch}{0.92}

    \begin{tabular}{p{0.4\textwidth}p{0.15\textwidth}p{0.15\textwidth}}
        \toprule
        \textbf{Name} & \textbf{Path Count} & \textbf{Percentage} \\
        \midrule
        Blackfriars Bridge & 59 & 2.22\% \\
        Hungerford Bridge and Golden Jubilee Bridges & 18 & 0.68\% \\
        Lambeth Bridge & 9 & 0.34\% \\
        London Bridge & 145 & 5.46\% \\
        Millennium Bridge & 37 & 1.39\% \\
        Southwark Bridge & 57 & 2.15\% \\
        Tower Bridge & 424 & 15.96\% \\
        Waterloo Bridge & 17 & 0.64\% \\
        Westminster Bridge & 67 & 2.52\% \\
        \midrule
        \textbf{Total} & 833 & 31.35\% \\
        \bottomrule
    \end{tabular}

    \vspace{0.4em}

    \includegraphics[width=\textwidth]{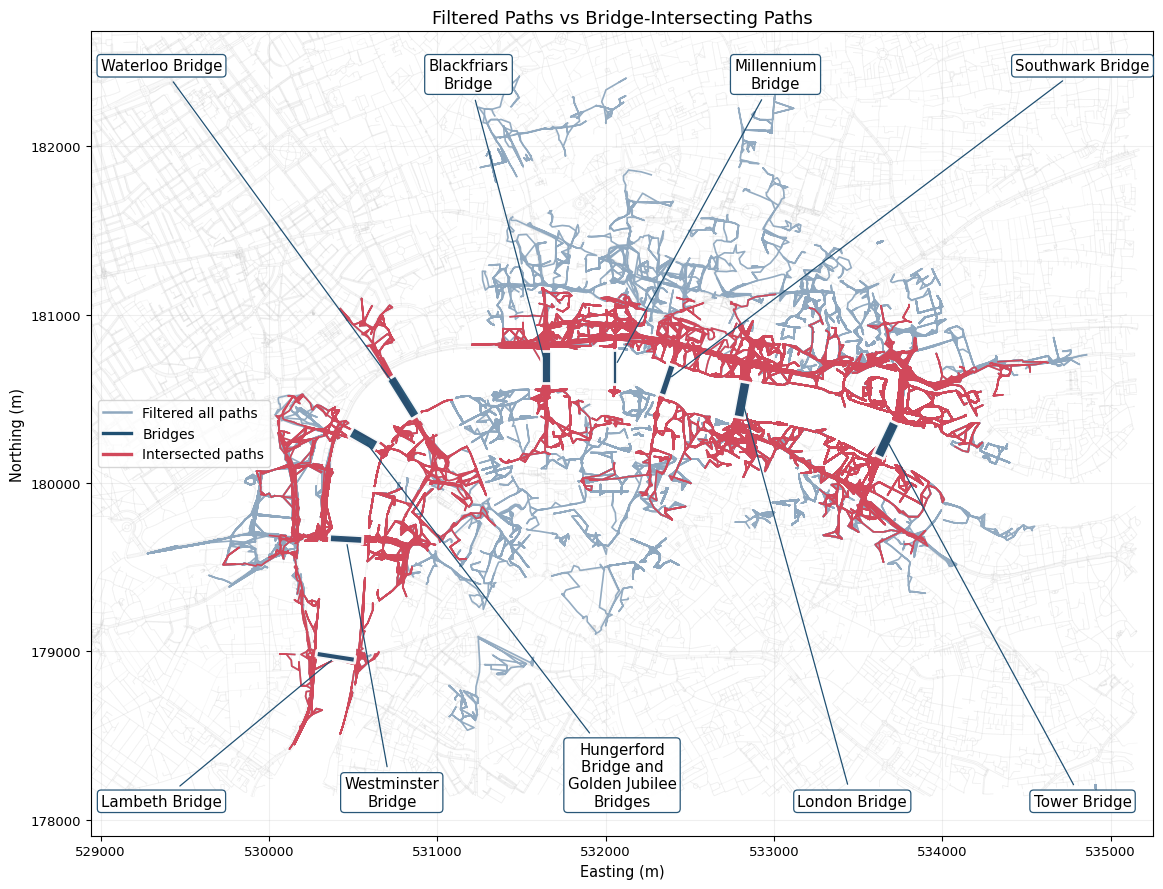}

    \vspace{-0.3em}

    \captionof{figure}{Distribution of VAV paths that pass over bridges and those that do not pass over bridges.}
    \label{fig:bridge_passing_paths}
\end{table}

\newpage
\bibliographystyle{elsarticle-harv}

\end{document}